\documentclass{article}

\usepackage{wrapfig}
\usepackage{CJKutf8} 
\usepackage{tabularx} 
\usepackage{longcat_style}
\usepackage{longtable}
\usepackage{xcolor}

\definecolor{tacticcolor}{rgb}{0.0, 0.1, 0.6}

\definecolor{codegreen}{rgb}{0,0.6,0}
\definecolor{codegray}{rgb}{0.5,0.5,0.5}
\definecolor{codepurple}{rgb}{0.58,0,0.82}
\definecolor{backcolour}{rgb}{0.91,0.91,0.9}
\definecolor{shallowRed}{rgb}{1, 0.8, 0.8}
\definecolor{shallowYellow}{rgb}{1, 0.953, 0.8}
\definecolor{shallowBlue}{rgb}{0.8, 0.8, 1}
\definecolor{orange}{rgb}{1, 0.6, 0}
\definecolor{shallowOrange}{rgb}{1, 0.88, 0.7}

\definecolor{keywordcolor}{rgb}{0.7, 0.1, 0.1}   
\definecolor{commentcolor}{rgb}{0.4, 0.4, 0.4}   
\definecolor{symbolcolor}{rgb}{0.0, 0.1, 0.6}    
\definecolor{sortcolor}{rgb}{0.1, 0.5, 0.1}      
\definecolor{errorcolor}{rgb}{1, 0, 0}           
\definecolor{stringcolor}{rgb}{0.5, 0.3, 0.2}    

\definecolor{leftcolor}{rgb}{0.522, 0.765, 0.863}  
\definecolor{midcolor}{rgb}{0.855, 0.776, 0.812}
\definecolor{rightcolor}{rgb}{0.886, 0.753, 0.596}

\usepackage[utf8]{inputenc} 
\usepackage[T1]{fontenc}    
\usepackage{newunicodechar}
\newunicodechar{，}{,}
\usepackage{hyperref}       
\usepackage{xcolor}
\usepackage{svg} 
\hypersetup{
    colorlinks=true,      
    linkcolor=blue,      
    urlcolor=blue,       
    citecolor=blue,      
    linkbordercolor=blue, 
    urlbordercolor=blue,
    citebordercolor=blue,
    pdfborderstyle={/S/U/W 1}, 
}

\usepackage{xurl}            
\usepackage{booktabs}       
\usepackage{amsfonts}       
\usepackage{nicefrac}       
\usepackage{microtype}      
\usepackage{lipsum}		
\usepackage{graphicx}
\usepackage{natbib}
\usepackage{doi}
\usepackage{amsmath}
\usepackage{amssymb} 
\usepackage{xspace}
\usepackage{enumitem}
\usepackage{multirow}
\usepackage{subcaption} 
\usepackage{makecell}
\usepackage{tcolorbox}
\usepackage{fancyvrb}
\usepackage{listings}
\tcbuselibrary{listings,skins,breakable}
\usepackage{hyperref, cleveref}
\usepackage{xcolor}

\usepackage{pmboxdraw}

\setlist[itemize]{leftmargin=*}
\setlist[enumerate]{leftmargin=*}
\setlist[description]{leftmargin=*}
\newcommand{\longcatthink}{LongCat-Flash-Thinking\xspace}
\newcommand{\longcatthinknew}{LongCat-Flash-Thinking-2601\xspace}
\newcommand{\longcatprover}{LongCat-Flash-Prover\xspace}

\def\lstlanguagefiles{./utils/lstlean.tex}

\lstdefinelanguage{json}{
    basicstyle=\ttfamily\small,
    backgroundcolor=\color{gray!12!white},
    frame=single,
    framerule=0.4pt,
    rulecolor=\color{gray!50!black},
    showstringspaces=false,
    breaklines=true,
    literate=
     *{0}{{{\color{gray!60!black}0}}}{1}
      {1}{{{\color{gray!60!black}1}}}{1}
      {:}{{{\color{gray!70!black}{:}}}}{1}
      {,}{{{\color{gray!70!black}{,}}}}{1}
      {\{}{{{\color{gray!80!black}{\{}}}}{1}
      {\}}{{{\color{gray!80!black}{\}}}}}{1},
    string=[s]{"}{"},
    stringstyle=\color{gray!80!black},
}

\usepackage{colortbl}
\definecolor{mygray}{gray}{.88}
\definecolor{mycyan}{cmyk}{.15,0,0,0}
\definecolor{mycyan2}{cmyk}{.85,0,0,0}

\definecolor{mygreen}{rgb}{0.19, 0.79, 0.02}
\definecolor{midnightgreen}{rgb}{0.0, 0.29, 0.33}

\title{\longcatprover: Advancing Native Formal Reasoning via Agentic Tool-Integrated Reinforcement Learning}

\author{ Meituan LongCat Team \\
	\texttt{longcat-team@meituan.com} \\
}

\renewcommand{\headeright}{\raisebox{-0.2\height}{\includegraphics[height=1.8em]{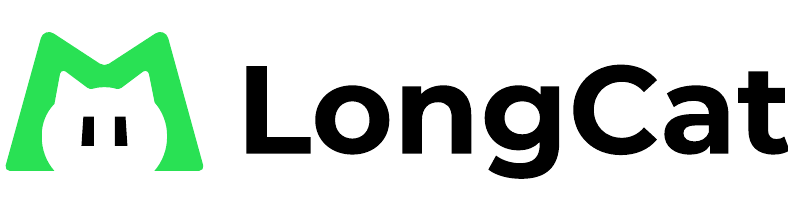}}}
\renewcommand{\shorttitle}{\longcatprover Technical Report}

\begin{document}
\maketitle

\begin{abstract}

We introduce~\textbf{\longcatprover}, a flagship $560$-billion-parameter open-source Mixture-of-Experts (MoE) model that advances \textbf{Native Formal Reasoning} in Lean4 through agentic tool-integrated reasoning (TIR).
We decompose the native formal reasoning task into three independent formal capabilities, i.e., \emph{auto-formalization}, \emph{sketching}, and \emph{proving}. 
To facilitate these capabilities, we propose a \textbf{Hybrid-Experts Iteration Framework} to expand high-quality task trajectories, including generating a formal statement based on a given informal problem, producing a whole-proof directly from the statement, or a lemma-style sketch.
During agentic RL, 
we present a \textbf{Hierarchical Importance Sampling Policy Optimization (HisPO)} algorithm, which aims to stabilize the MoE model training on such long-horizon tasks. It employs a gradient masking strategy that accounts for the policy staleness and the inherent train-inference engine discrepancies at both sequence and token levels.
Additionally, we also incorporate theorem consistency and legality detection mechanisms to eliminate reward hacking issues.
Extensive evaluations show that our~{\longcatprover} sets a new state-of-the-art for open-weights models in both auto-formalization and theorem proving. Demonstrating remarkable sample efficiency, it achieves a 97.1\% pass rate on MiniF2F-Test using only 72 inference budget per problem. On more challenging benchmarks, it solves 70.8\% of ProverBench and 41.5\% of PutnamBench with no more than 220 attempts per problem, significantly outperforming existing open-weights baselines.


\textbf{Huggingface}: \href{https://huggingface.co/meituan-longcat/LongCat-Flash-Prover}{https://huggingface.co/meituan-longcat/LongCat-Flash-Prover}\\
\textbf{Project Page}: \href{https://github.com/meituan-longcat/LongCat-Flash-Prover}{https://github.com/meituan-longcat/LongCat-Flash-Prover}\\
\end{abstract}


\begin{figure}[hb]
    \centering
    \begin{subfigure}{0.282\textwidth}
        \centering
        \includegraphics[width=\linewidth]{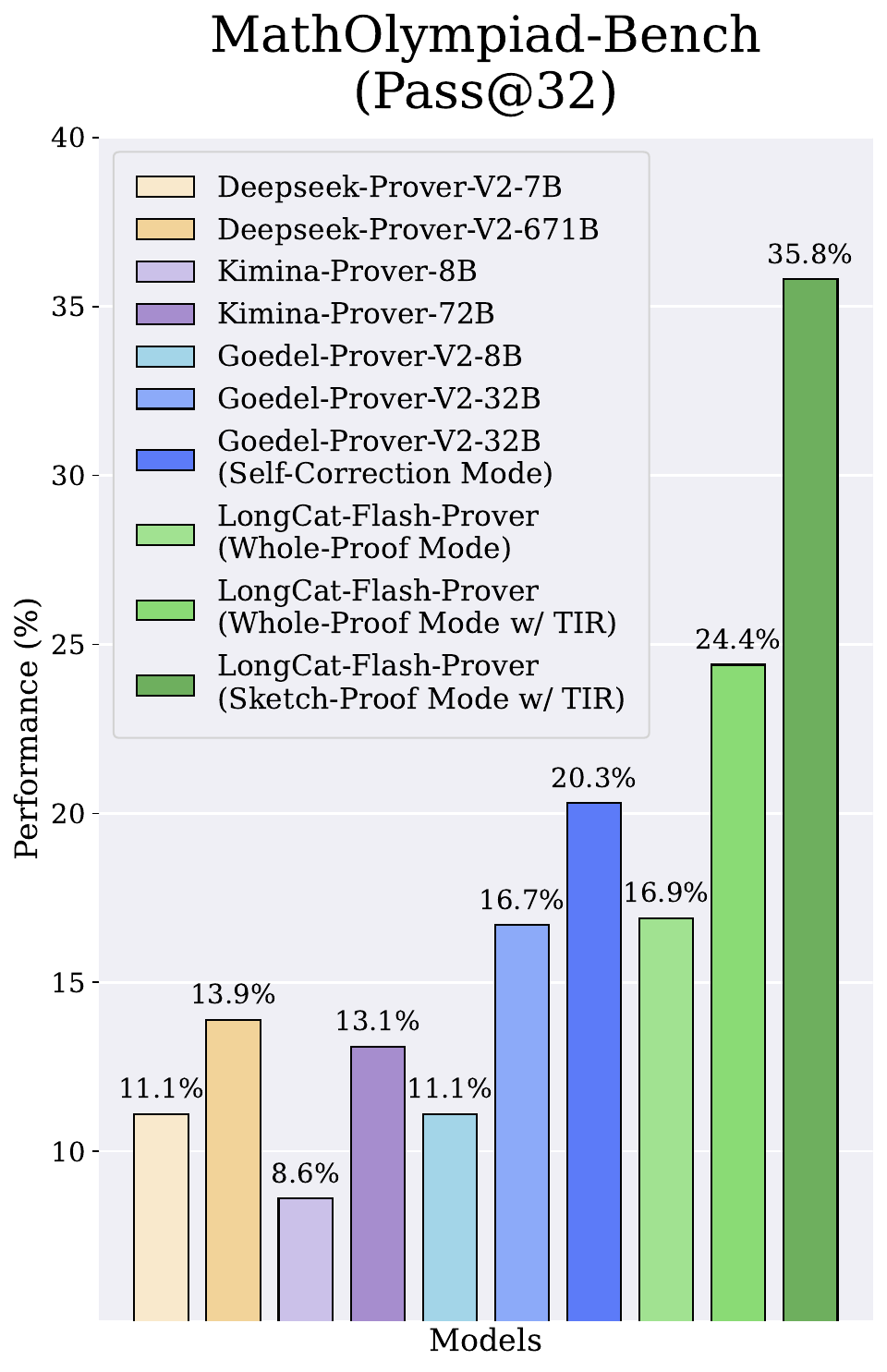}
        \label{fig:matholympiad_bench}
    \end{subfigure}
    \hfill 
    \begin{subfigure}{0.282\textwidth}
        \centering
        \includegraphics[width=\linewidth]{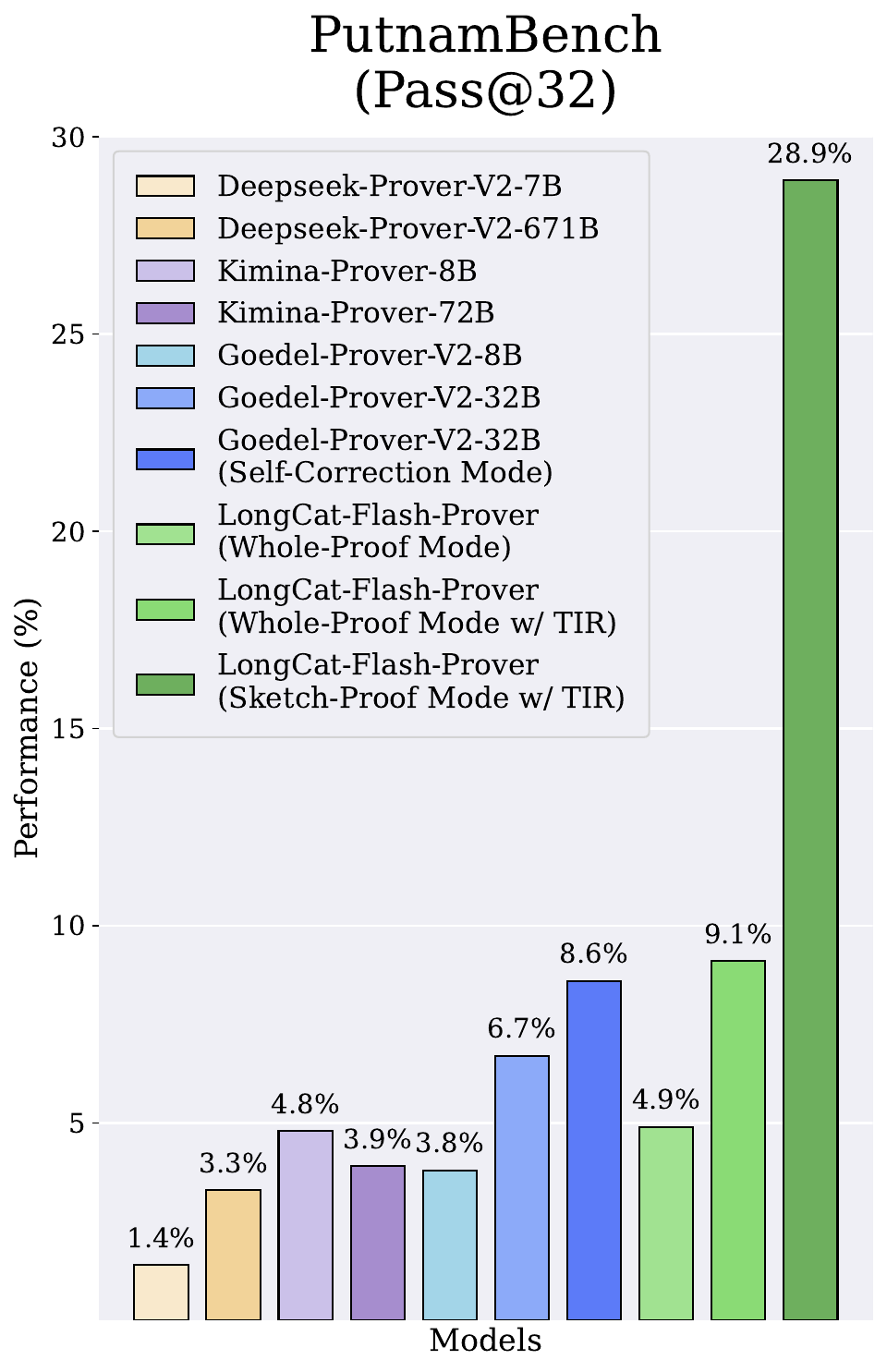}
        \label{fig:putnambench}
    \end{subfigure}
    \hfill
    \begin{subfigure}{0.406\textwidth}
        \centering
        \includegraphics[width=\linewidth]{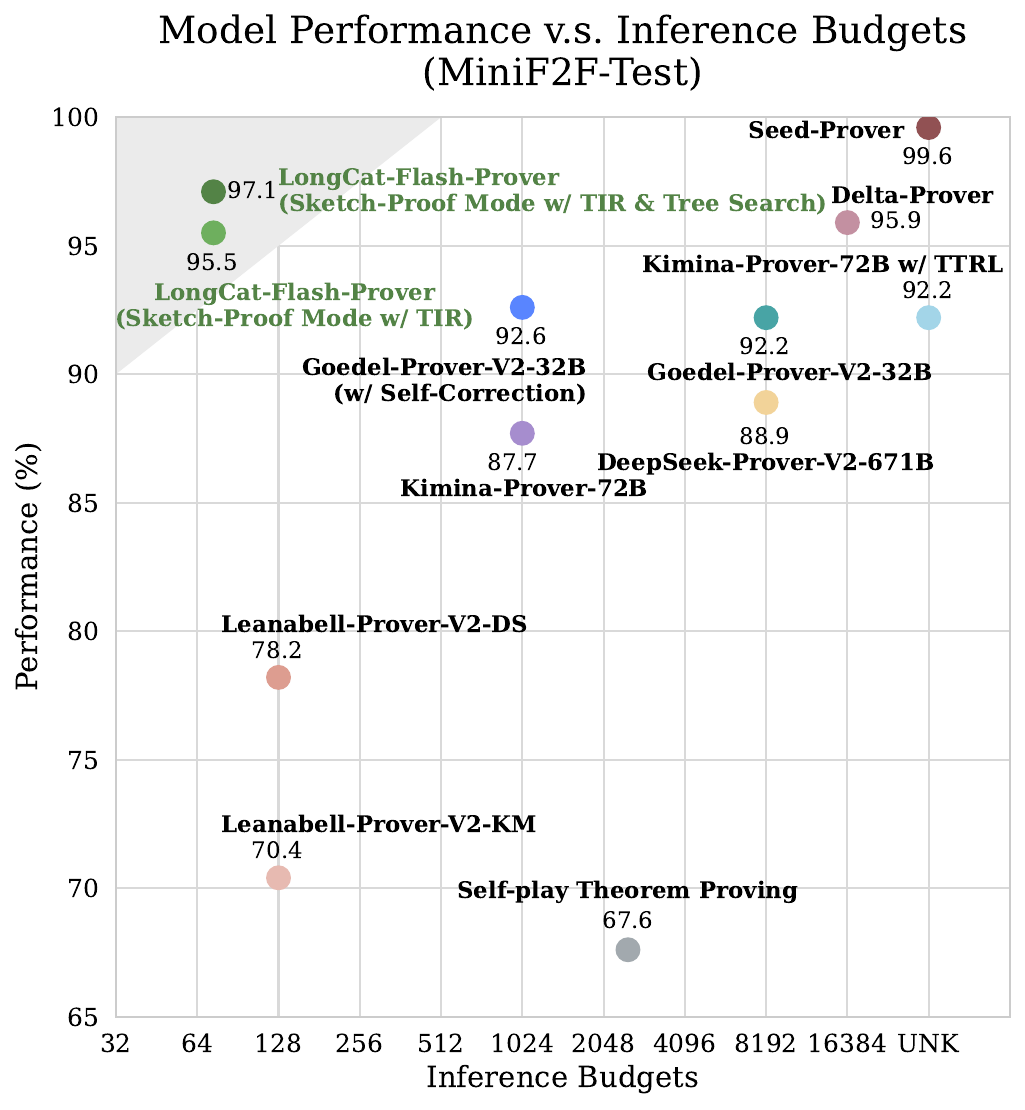}
        \label{fig:minif2f_test_vs_budget}
    \end{subfigure}
    
    \caption{The performance comparison over proving tasks. The figures on the left and middle illustrate the performance with a limited 32 inference budget on MathOlympiad-Bench and PutnamBench, respectively. The figure on the right shows the model performance on MiniF2F-Test balancing with the inference budgets.}
    \label{fig:head_result}
\end{figure}


\newpage

\section{Introduction}

Recent advancements in large language models (LLMs) have shifted decisively toward enriching the reasoning capabilities, promoting the boundaries of artificial general intelligence (AGI)~\citep{openai_o1, comanici2025gemini, guo2025deepseek, yang2025qwen3}.
While notable progress has been made in solving complex problems using natural language, current LLMs still struggle with formal theorem-proving tasks. These tasks necessitate the use of rigorous, verified formal languages (e.g., Lean4) to ensure reliable formal statements and proofs.
Several previous efforts have been devoted to leveraging feedback from verification tools to train models in repairing Lean4 code snippets~\citep{shang2025stepfunproverpreviewletsthink, wang2025kiminaproverpreviewlargeformal, lin2025goedelproverv2scalingformaltheorem, chen2025seedproverdeepbroadreasoning, chen2025seedprover15masteringundergraduatelevel, ji2025leanabellproverv2verifierintegratedreasoningformal, shen2025realproverretrievalaugmentedlean, lin2025leanstarlearninginterleavethinking, xin2024deepseekproverv15harnessingproofassistant, ren2025deepseekproverv2advancingformalmathematical}.
However, unlike traditional Python scripts or other callable tools, Lean4 is a formal language that embodies the rigorous logical progression of a solution. Consequently, directly applying vanilla TIR to such formal verification tasks remains a significant challenge.

In this work, we introduce {\longcatprover}, an efficient open-source Mixture-of-Experts (MoE) reasoning model that sets a new state-of-the-art for open-source reasoning models. 
{\longcatprover} is built upon our foundational LongCat Mid-train Base model~\citep{meituan2025longcat_flash_chat}, which comprises 560B total parameters with approximately 27B active parameters. 
Compared to our recent {\longcatthinknew}~\citep{meituanlongcatteam2026longcatflashthinking2601technicalreport} and other open-source models~\citep{meituan2025longcat_flash_chat, meituanlongcatteam2025introducinglongcatflashthinkingtechnicalreport}, {\longcatprover} represents a significant leap in both informal reasoning (e.g., logic, mathematics, coding, and agentic tasks) and formal reasoning (e.g., auto-formalization and theorem proving).
The development of {\longcatprover} is characterized by the following key innovations:
\begin{itemize}
    \item \textbf{Native formal reasoning}. We define ``native formal reasoning'' as a core capability of LLMs, analogous to native multimodal~\citep{shukor2025scalinglawsnativemultimodal} and native tool calls~\citep{anthropic_claude_opus_4_6}. This paradigm enables the model to leverage formal operators to solve complex reasoning tasks without specialized architectural modifications.
    We decompose the native formal reasoning into 
    three specific capabilities: 1) \textbf{Agentic auto-formalization} aims to transform the informal statement~\footnote{The problem or reasoning trajectory that in natural language can be denoted as informal statement.} into a verified formal statement; 2) \textbf{Agentic sketching} aims to generate a lemma-style sketch based upon the given problem and corresponding formal statement~\citep{Jiang2023Draft}; 3) \textbf{Agentic proving} aims to generate a whole-proof that completes the target theorem body, or to generate a lemma-style proof that introduces helper lemmas and finally proves the target theorem. These capabilities are further enhanced through a TIR strategy, where all experts can interact directly with the Lean4 tools for compilation and verification.
    
    \item \textbf{Hybrid-experts iteration framework}. To facilitate native formal reasoning, we developed a framework to generate high-quality cold-start data. This framework employs several optimized expert models,  each specialized in distinct domains such as auto-formalization, lemma-style sketching, and proving. 
    We utilize this framework to synthesize a series of trajectories centered on native formal operators, using multiple verifiable formal tools as environmental feedback.
    By doing so, each expert is iteratively refined on these tool-assisted reasoning trajectories, emulating the human process of learning through trial, verification, and reflection.

    \item \textbf{Hierarchical Importance Sampling Policy Optimization (HisPO)}.
    Following our prior works~\citep{meituanlongcatteam2026longcatflashthinking2601technicalreport, meituanlongcatteam2025introducinglongcatflashthinkingtechnicalreport}, we perform agentic reinforcement learning with verified reward (RLVR) by designing different tasks, including generating a formal statement based on a given informal problem, producing a proof directly from the statement, or a lemma-style sketch. 
    To make the MoE model training stable, we introduce HisPO, which is a hierarchical clipping strategy that eliminates the gradient contributions who has large training-inference engine discrepancy by estimating sequence-wise or token-wise important sampling (IS) ratios. 
    In addition to outcome-based rewards, we designed a legality detection strategy to explore the proof with obvious hacking features, for example, the proof that is inconsistent with the semantics of the formal statement, mismatching the pre-defined theorem conditions, containing unverified or model-created axioms that attempt to fool the Lean4 server, etc.

    
    

\end{itemize}

We conduct extensive experiments on multiple challenging benchmarks to evaluate the effectiveness of our~{\longcatprover} on solving native formal reasoning tasks.
As shown in Figure~\ref{fig:head_result},
Compared to current open-source state-of-the-art model, our model with TIR strategy achieves a significant improvement of 25.5\% and 20.3\% on MathOlympiad-Bench and PutnamBench in the Pass@32 metric, respectively. 
Furthermore, by increasing the inference budget, we achieved a score of 97.1\% on the MiniF2F-Test with only 72 attempts per problem, surpassing the performance and efficiency of open-source SOTA models.
In addition, our model outperforms all open-source general-purpose reasoning models and proprietary models on auto-formalization and other proving tasks, and maintains a performance level comparable to~{\longcatthinknew} on the informal reasoning task.
We hope our open-source native formal reasoning model will encourage the community to build upon it, prompting AI systems to better solve complex problems in both informal and formal scenarios.

\section{Hybrid-Experts Iteration Framework}

\begin{figure}
\centerline{\includegraphics[width=0.95\linewidth]{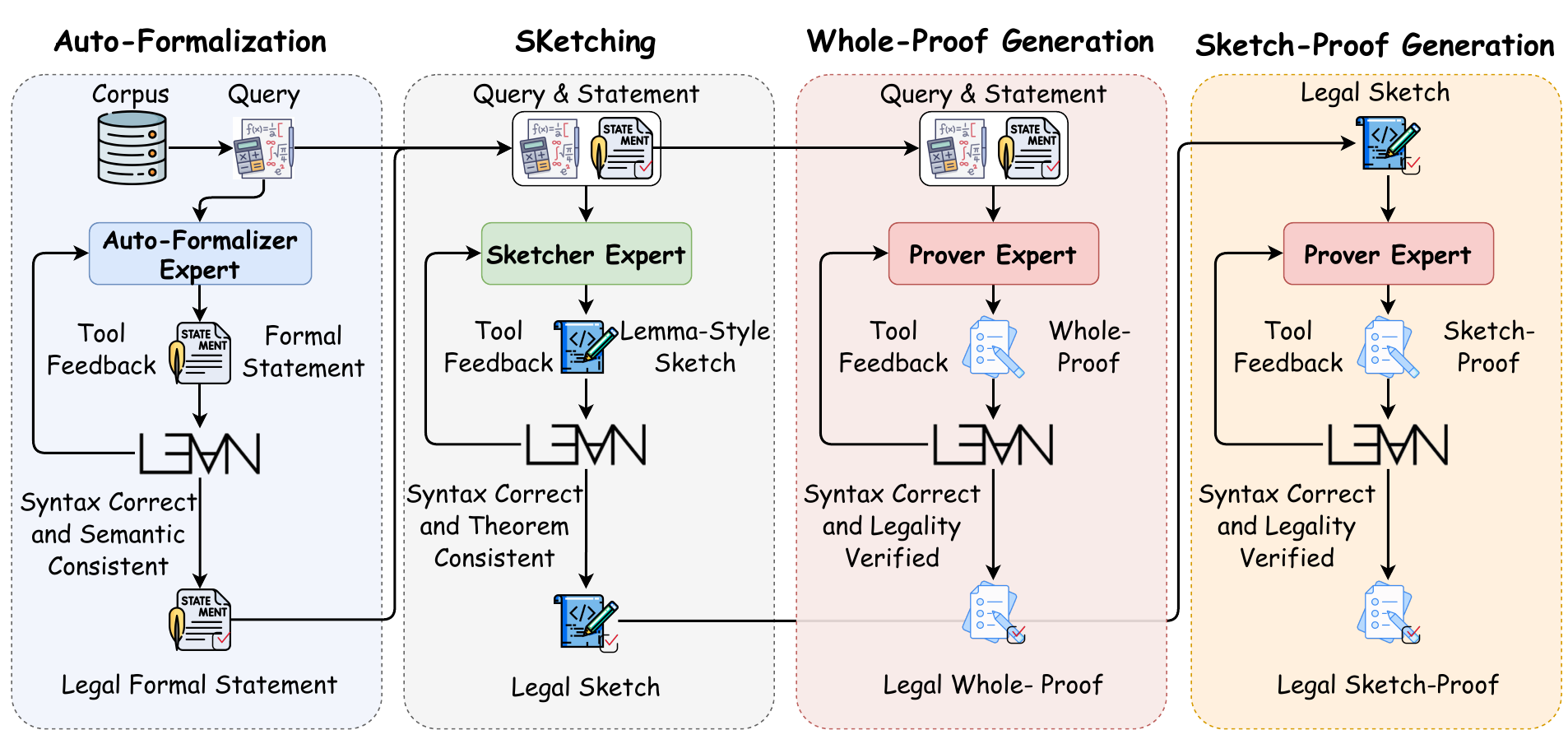}}
\caption{\small
The overview of our 
hybrid-experts tool-integration synthesis pipeline. 
In this pipeline, we iteratively optimize three different experts (i.e., auto-formalizer, lemma-style sketcher, and prover), and use these experts to synthesize trajectories based on pre-defined formal reasoning capabilities.
Given one problem in natural language, we first transform it into a formal statement in Lean4 by interacting with the Lean4 compiler.
Then, the formal statement will be used to generate a whole-proof and lemma-style sketch. 
If the whole-proof still fails to pass verification within a limited number of tool feedback rounds, the proof generated from the lemma-style sketch will be used instead for verification.
In the rejection sampling phase, we will retain the trajectory that allows us to fully utilize the tools to perform auto-formalization, sketching, and proving.
}
\label{fig:expert_iteration}
\end{figure}

To help realize this potential and empower researchers, we introduce significant enhancements to our model's formal reasoning capabilities. 
Our work aims to provide a robust and versatile foundation upon which the community can build and explore new scientific frontiers.
To reach this goal, we develop a hybrid-experts iteration framework to synthesize multiple verified trajectories and continuously train professional native formal reasoning expert models. 

\subsection{Native Formal Experts }

We decompose native formal reasoning into three main capabilities: \emph{auto-formalization}, \emph{sketching}, and \emph{proving}. Each capability corresponds to a specialized expert model,
formally, representing $\pi_{\theta_{af}}$, $\pi_{\theta_{sk}}$, and $\pi_{\theta_{pf}}$, respectively.
We also maintain a set of available tools $\mathcal{T}_{af}$, $\mathcal{T}_{sk}$, and $\mathcal{T}_{pf}$ to work with the expert model for the TIR trajectories synthesis and rejected sampling.

\noindent\paragraph{Auto-Formalization (AF)}
The task of AF aims to transform the natural language problem statement or unfinished proof into a formal statement in Lean4.
Formally, the input denoted as $x$ is uniformly referred to as an informal statement.
Through the AF, we can obtain a formal statement as $s_x = \pi_{\theta_{af}}(x)$.
We found that the model outputs syntactically incorrect or semantically inconsistent formal statements during the AF process. 
To address this, we introduce two verified tools to form the tool set $\mathcal{T}_{af}=\{\mathcal{V}_{syn}, \mathcal{V}_{con}\}$.
\begin{itemize}
    \item \textbf{Statement Syntax Detection $\mathcal{V}_{syn}$}: We follow~\citep{wang2025kimina}'s work to develop the Lean4 Server~\footnote{\url{https://github.com/project-numina/kimina-lean-server}.} (v4.15). Each generated formal statement is concatenated with the placeholder ``:= by sorry'', and compiled through the Lean4 Server. 
    This tool yield a binary outcome as $\mathcal{V}_{syn}(s_x)\in\{\text{SORRY}, \text{FAIL}\}$, where $\text{SORRY}$ means the statement has no syntax error other than the ``:= by sorry'' part not being implemented.
    In default, we direct receive the JSON-like feedback from this tool, which consists of the error messages and the location. 
    \item \textbf{Semantic Consistency Detection $\mathcal{V}_{con}$}: We found that auto-formalization can occasionally alter the original problem's meaning.
    To address this, we employ a model-based semantic filter to identify and discard formal statements that are inconsistent with their informal counterparts. The specific implementation is shown in Appendix~\ref{appendix:consistency_prompt}.
\end{itemize}

\noindent\paragraph{Sketching}
The task of sketching aims to generate a lemma-style sketch in Lean4 for a given formal statement. A lemma-style sketch resembles a functional programming approach, consisting of a complete proof body for the target theorem, following several unproved helper lemmas (initially admitted ``:= by sorry''). Introducing the sketch strategy is inspired by Divide and Conquer as well as Dynamic Programming, since the helper lemmas can be much easier to prove under appropriate decomposition and proved helper lemmas can be referenced in proof of following lemmas. Formally, given a natural language problem $x$ and the corresponding verified formal statement $s_x$. The sketch can be represented as $d_x=\pi_{\theta_{sk}}(x, s_x)$, where $d_x$ contains $n$ helper lemmas and the main body, i.e., $d_x=[lemma_1, \cdots, lemma_n, s_x, body_x]$, $lemma_i$ denotes the $i$-th helper lemma, and $body_x$ denotes the main proof body of the target theorem.
A case of the lemma-style sketch is shown in Appendix~\ref{apdx:sketch_case}.

\noindent\paragraph{Proving}
Formally, the task of theorem proving is to generate a valid proof in Lean4. We define two kinds of proving schema according to the input format.
\begin{itemize}
    \item \textbf{Whole-Proof Generation}: Given a natural language problem $x$ and the corresponding formal statement $s_x$, the task of whole-proof generation is to produce a proof $p_x$ in a single interaction (without tools) or multi-interaction (with tools). Formally, we have $p_x=\pi_{\theta_{pf}}(x, s_x)$.
    \item \textbf{Sketch-Proof Generation}: Given a natural language problem $x$ and the corresponding formal statement $s_x$, we first apply the sketcher model to construct a formal sketch $d_x$, where the target theorem is proved on the basis of several unproved helper lemmas. Then we let the prover model complete the proofs of unproved lemmas, so as to produce a final proof. Formally, we have $p_x=\pi_{\theta_{pf}}(x, d_x)$.
\end{itemize}
For the tools, we design two verifiers $\mathcal{T}_{pf}=\{\mathcal{V}_{syn}, \mathcal{V}_{val}\}$ to detect whether the proof is valid:
\begin{itemize}
    \item \textbf{Syntax Verificiation $\mathcal{V}_{syn}$}: This tool aims to check the proof syntax, yielding an outcome from $\mathcal{V}_{syn}(p_x)\in\{\text{SORRY}, \text{PASS}, \text{FAIL}\}$. The PASS outcome means that the current proof is successfully compiled by the Lean4 kernel, without any formal errors or unproved statements, which would lead to FAIL and SORRY respectively.
    \item \textbf{Legality Detection $\mathcal{V}_{leg}$}: A formally complete Lean4 code may be inconsistent with the original formal statement, e.g., with tampered theorem definition or special compilation context. To this end, we develop light-weight Lean4 lexer and parser to convert Lean4 code into Abstract Syntax Tree (AST), and perform strict AST consistency checks between the formal statement and the proof or the sketch. The details are shown in Appendix~\ref{app:legality}.
\end{itemize}

\subsection{Hybrid-experts Tool-Integration Synthesize}
\label{sec:curriculum}

As shown in Figure~\ref{fig:expert_iteration}, we utilize these expert models for data synthesis. Each expert model generates either single-turn trajectories (without tool calls) or multi-turn trajectories (in TIR mode), thereby ensuring diversity in the synthesized data. 
To enable the model to dynamically select appropriate tools and proof strategies based on the difficulty of each problem, we adopt a curriculum learning approach: 1) starting with single-turn synthesis, then followed by multi-turn tool calls synthesis, and 2) progressing from whole-proof generation to lemma-style sketch proving. 
The basic synthesis process is as follows:
\begin{enumerate}
    \item Give a natural language problem $x_i$, we first use the auto-formalizer $\pi_{\theta_{af}}$ to generate $N$ responses. For each response, we extract the formal statement in Lean4, and perform rejected sampling via the pre-defined tools $\mathcal{T}_{af}$. If at least one formal statement passes validation, we retain all verified statements and construct the AF single-turn trajectory set, represented as:
\begin{equation}
\begin{aligned}
\mathcal{D}_{af}=\{(x_i, s_{x_i}|\mathcal{V}_{syn}(s_{x_i})=\text{SORRY}\cap\mathcal{V}_{con}(s_{x_i})=1\}_i^{\leq N}.
\end{aligned}
\end{equation}

    \item If $x_i$ does not correspond to the correct formal statement within $N$ attempts, we will use the TIR mode of auto-formalization. Specifically, we select the response with the fewest tool feedback errors from $N$ trajectories as the first turn, and then let the auto-formalizer re-think based on the tool feedback until it finally passes the verification of all tools.
    Formally, the synthesized trajectories can be formed as:
\begin{equation}
\begin{aligned}
\mathcal{D}'_{af}=\{(x_i, s_{x_i}^{1}, \tau_{s_{x_i}}^{1}, \cdots, s_{x_i}^{m}|\mathcal{V}_{syn}(s_{x_i}^{m})=\text{SORRY}\cap\mathcal{V}_{con
    }(s_{x_i}^{m})=1\}_i^{\leq N},
\end{aligned}
\end{equation}
     where $\tau_{s_{x_i}}^{j}=\mathcal{V}_{syn}(s_{x_i}^{j})+\mathcal{V}_{con}(s_{x_i}^{j})$ is the tool feedback at $j$-th interaction, $m$ denotes to the model-tool interaction times, we will denote $s_{x_i}=s_{x_i}^{m}$ as the final formal statement.

    \item Next, for each problem $x_i$ that has at least one verified formal statement, we will use the prover model $\pi_{\theta_{pf}}$ to perform whole-proof generation.
    We randomly sample one formal statement $s_x$, and let the prover model generate $N$ whole-proofs.
    Similar to the synthesis strategy of auto-formalization, we retain all correct proofs to reach a final trajectory set:
\begin{equation}
\begin{aligned}
\mathcal{D}_{whole.pf} = \{(x_i, s_{x_i}, p_{x_i})| (x_i, s_{x_i})\in\mathcal{D}_{af}\cup\mathcal{D}'_{af}, \mathcal{V}_{syn}(p_{x_i})=\text{PASS}\cap\mathcal{V}_{leg
    }(p_{x_i})=1\}_i^{\leq N}\}.
\end{aligned}
\end{equation}

    \item For problems that fail after $N$ attempts at whole-proof generation, we select the response with the fewest errors and leverage tool feedback to enable the prover model to interact with verification tools. Steps 3 and 4 are repeated multiple times, each time using a different formal statement as input to the prover. This process is designed to ensure that the same problem is associated with diverse formal statements and proof trajectories, thereby increasing the likelihood of discovering a correct proof, regardless of whether TIR is employed. Finally, the trajectories set of whole-proof with TIR refers to:
\begin{equation}
\begin{aligned}
\mathcal{D}'_{whole.pf} = \{(x_i, s_{x_i}, p_{x_i}^{1}, \tau_{p_{x_i}}^{1}, \cdots, p_{x_i}^{m})| (x_i, s_{x_i})\in\mathcal{D}_{af}\cup\mathcal{D}'_{af}, \mathcal{V}_{syn}(p_{x_i}^{m})=\text{PASS}\cap\mathcal{V}_{leg
    }(p_{x_i}^{m})=1\}_i^{\leq N}\},
\end{aligned}
\end{equation}
    where $\tau_{p_{x_i}}^{j}=\mathcal{V}_{syn}(p_{x_i}^{j})+\mathcal{V}_{leg}(p_{x_i}^{j})$ is the tool feedback at $j$-th interaction, we denote $p_{x_i}=p_{x_i}^{m}$ as the final proof.

    \item For unresolved problems that have a verified formal statement yet lack a verified proof, we deploy the sketcher model to generate $N$ lemma-style sketches. This decomposition into multiple lemmas is designed to ease and enhance the efficiency of subsequent proof construction.
    Specifically, we aim to produce $N$ responses and extract the Lean4 sketch to perform rejected sampling. The sketch will be retained, which contains the SORRY result and a consistent theorem. We directly use TIR mode to increase the likelihood of discovering the verified sketch:
\begin{equation}
\begin{aligned}
\mathcal{D}'_{sk} = \{(x_i, s_{x_i}, d_{x_i}^{1}, \tau_{d_{x_i}}^{1}, \cdots, d_{x_i}^{m})| (x_i, s_{x_i})\in\mathcal{D}_{af}\cup\mathcal{D}'_{af}, \mathcal{V}_{syn}(d_{x_i}^{m})=\text{SORRY}\cap\mathcal{V}_{theo
    }(d_{x_i}^{m})=1\}_i^{\leq N}\},
\end{aligned}
\end{equation}
where $\tau_{d_{x_i}}^{j}=\mathcal{V}_{syn}(d_{x_i}^{j})+\mathcal{V}_{theo}(d_{x_i}^{j})$ is the tool feedback at $j$-th interaction, the final sketch is $d_{x_i}=d_{x_i}^{m}$.

    \item Through the lemma-style sketch $d_{x_i}$, we can reuse the prover model to complete each lemma, which is a similar action to generate whole-proof for each lemma. We directly use TIR mode to synthesize the lemma-style proof:
\begin{equation}
\begin{aligned}
\mathcal{D}'_{sk.pf} = \{(x_i, d_{x_i}, p_{x_i}^{1}, \tau_{p_{x_i}}^{1}, \cdots, p_{x_i}^{m})| (x_i, d_{x_i})\in\mathcal{D}_{sk}\cup\mathcal{D}'_{sk}, \mathcal{V}_{syn}(p_{x_i}^{m})=\text{PASS}\cap\mathcal{V}_{leg
    }(p_{x_i}^{m})=1\}_i^{\leq N}\}.
\end{aligned}
\end{equation}

\end{enumerate}

To this end, the hybrid experts can reach 6 different sets of trajectories $\mathcal{D}_{af}, \mathcal{D}'_{af}, \mathcal{D}_{whole.pf}, \mathcal{D}'_{whole.pf}, \mathcal{D}'_{sk}, \mathcal{D}'_{sk.pf}$.
Importantly, within this framework, single-turn trajectories (e.g., $\mathcal{D}_{af}, \mathcal{D}_{whole.pf}$) that do not require tool interaction typically indicate a relatively simple task.
Trajectories requiring tool interaction (e.g., $\mathcal{D}'_{af}, \mathcal{D}'_{whole.pf}, \mathcal{D}'_{sk}, \mathcal{D}'_{sk.pf}$) indicate a more difficult task. 
This progressive synthesis approach (first synthesizing without tools, then synthesizing with tool feedback) allows the model to dynamically perceive the task's difficulty and its adaptability to tool invocation.
The specific cases of different types of trajectories are illustrated in Appendix~\ref{app:cases}.

\subsection{Experts Self-Evolving}

To enrich the trajectories and the experts, we follow an expert iterative strategy.
Each iteration begins with a cold-start process that progressively refining through self-distillation, and is followed by an agentic RL process.
For this purpose, we utilize our prior LongCat Mid-train Base Model as the foundation for the auto-formalizer, prover, and sketcher.

Through this iterative process, we curated a substantial corpus of high-quality training instances, each containing a formal statement, a synthesized thinking process, and a verified proof. 
This dataset was then used to comprehensively enhance the native formal reasoning capabilities of our~{\longcatthinknew}.
We will describe the details in Section~\ref{sec:approach}.

\section{Approach}
\label{sec:approach}

\subsection{Overview}

\begin{figure}
    \centering
    \includegraphics[width=0.95\linewidth]{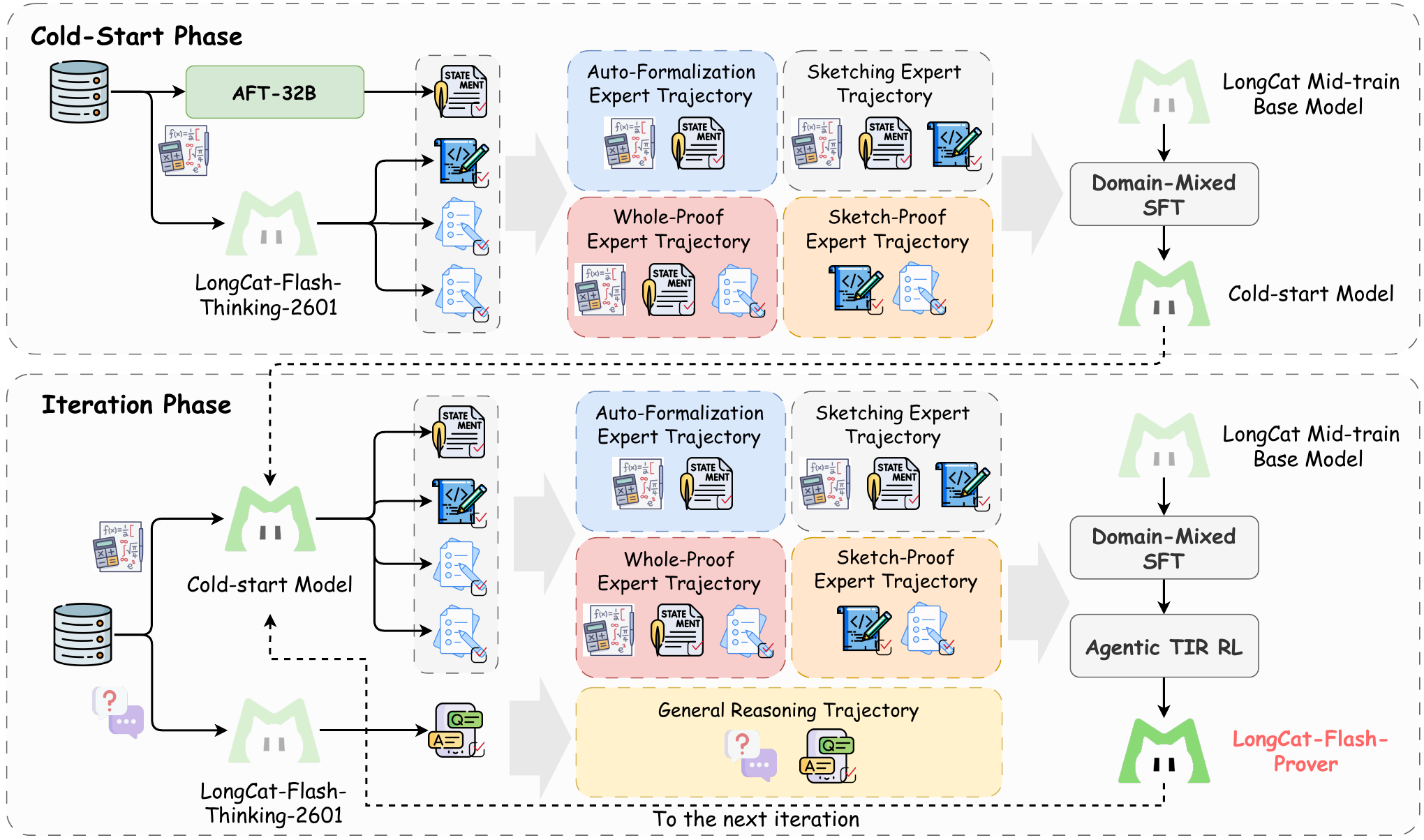}
    \caption{
    The overview of the training pipeline of~{\longcatprover}.
    We choose the LongCat Mid-train base model as the starting point, and then perform domain-mixed SFT to obtain a cold-start model. 
    Then, we use the new cold-start model to refresh the prompts and trajectories through self-distillation and agentic RL, and repeat this process multiple times. After the iteration, we obtain our~{\longcatprover} model.
    }
    \label{fig:training_pipeline}
\end{figure}

In this section, we describe the training approach for our {\longcatprover}.
The overview of the training process is shown in Figure~\ref{fig:training_pipeline}, it begins with an initial checkpoint derived from the LongCat Mid-train Base model, an early-stage version of our previous {\longcatthinknew}. 
The pipeline is organized into two main phases:
\begin{itemize}
    \item \textbf{Cold-start Phase}. We first utilize ATF-32B~\citep{guo2025autoformalizertoolfeedback}, our previously released auto-formalization model trained via TIR and DPO~\cite{DPO_nips}, to synthesize multiple formal statements. Building on these statements, we leverage our~{\longcatthinknew}~\citep{meituanlongcatteam2026longcatflashthinking2601technicalreport} to generate high-quality agentic trajectories integrated with verification tools. From this synthesized data, we curate a high-quality cold-start dataset by performing decontamination, deduplication, and sampling based on both difficulty and diversity. 
    Since different expert models come from different model families, we apply domain-mixed SFT to integrate these capabilities. Specifically, we employ the LongCat Mid-train Base model~\citep{meituanlongcatteam2026longcatflashthinking2601technicalreport} as the initialization point, and the trained model will be represented as the cold-start model.

    \item \textbf{Iteration Phase}. In the iteration phase, we choose the cold-start model derived from the cold-start phase as our new expert. 
    The trajectories for each native formal reasoning task are synthesized using this new expert. 
    In addition, we also incorporate a large amount of general data to ensure the model possesses informal reasoning capabilities. 
    After data curation, we train the model using domain-mixed SFT and agentic TIR RL, and further improve its performance via multiple iterations.
    Following the expert iteration cycles, we perform a final round of SFT and agentic TIR RL to reach our final~{\longcatprover}.

\end{itemize}




\subsection{Data Curation}
We curate our training data from two views, i.e., mathematics queries for native formal reasoning, and general-purpose queries for informal reasoning.
We mix up generic data in the cold start phase means preserving the informal reasoning capabilities of the~{\longcatprover} model in regular dialogues, including conversation, STEM-like reasoning, coding, agentic tool use, searching, knowledge, and safety.
We directly sample some pieces of cold-start data that are used by~{\longcatthinknew}, enabling our~{\longcatprover} to achieve comparable informal reasoning performance with~{\longcatthinknew}.

For native formal reasoning, we collect a variety of mathematical and formal queries from open-source datasets and in-house corpus. 
To further enhance data quality, we additionally developed an external self-synthesis framework that constructs complex problems by transforming informal tasks into formal ones, spanning auto-formalization, proving, and sketching. 

To make the hybrid-experts iteration framework efficiency by using these prompts, we built a simple data processing pipeline in each iteration process, including basic processing, difficulty estimation, and diversity sampling.

\noindent\paragraph{Basic Processing}

We adopt the data processing workflow of~{\longcatthinknew} for basic data preprocessing, which includes semantic deduplication, desaturation, and quality assurance checks.
Specifically, for natural language problems which are static in nature, preprocessing is conducted only at the initial stage of the expert iteration. 
In each expert iteration, we additionally preprocess the formal statements generated through the auto-formalizer expert, primarily by removing duplicates and filtering out statements that may closely resemble those in the test set. 
Unlike our previous mention involving rejection sampling with predefined tools, the formal statements in this context are inherently correct and verifiable; filtering is performed solely to prevent data leakage and ensure the integrity of the evaluation.

\noindent\paragraph{Difficulty Estimation}
As previously described in Section~\ref{sec:curriculum}, our hybrid-experts iterative framework incorporates the difficulty estimation of each prompt. 
Specifically, for each prompt, the expert will repeatedly synthesize into $N$
trajectories, allowing us to compute their difficulty using the following metric:
\begin{equation}
\begin{aligned}
\text{Difficulty}(x_i, \mathcal{D}) = \frac{\sum_{(x_j, \cdots)\in\mathcal{D}}\mathbb{I}(x_i=x_j)}{N},
\end{aligned}
\end{equation}
where $\mathcal{D}\in\{\mathcal{D}_{af}, \mathcal{D}'_{af}, \mathcal{D}_{whole.pf}, \mathcal{D}'_{whole.pf}, \mathcal{D}'_{sk}, \mathcal{D}'_{sk.pf}\}$ is the set that contains only verified trajectories.

This enables us, on one hand, to monitor the progress and evolution of each expert throughout the iterations, and on the other hand, to dynamically select more challenging data for subsequent training rounds. Prompts with a difficulty score of 0 are retained for future synthesis cycles, while those with a difficulty score of 1 for two or more consecutive iterations are removed from the synthesis process to enhance efficiency. 
Additionally, for each set involving tool calls (i.e., $\mathcal{D}'_{af}, \mathcal{D}'_{whole.pf}, \mathcal{D}'_{sk}, \mathcal{D}'_{sk.pf}$), it is further divided into two subsets for cold-start and RL training. 
Prompts allocated to cold-start training and RL training are both ensured to have a non-zero and non-one pass rate, respectively.

\noindent\paragraph{Diversity Sampling}
The expert iteration framework produces 6 distinct sets of synthesized trajectories. 
To prevent overfitting during cold-start training, we restrict each prompt to a single trajectory, facilitated by a diversity sampling strategy. We evaluate the average trajectory length and tool call frequency within each set, preferentially selecting responses with shorter lengths or fewer tool calls to avoid excessive reasoning or irrelevant tool usage.

Rather than simply choosing the shortest trajectory or the one with minimal tool calls, we implement a weighted sampling scheme. This approach ensures that each prompt is associated with only one trajectory, while preserving the diversity of the overall trajectory collection.

\subsection{Hierarchical Importance Sampling Policy Optimization}

For agentic RL, 
We introduce a novel HisPO algorithm that implements a hierarchical gradient masking strategy by estimating the training-inference consistency of sequence-level and token-level granularity to ensure a stable optimization.
We built this training recipe on our Dynamic ORchestration for Asynchronous rollout (DORA) system, which is introduced in our prior works~\citep{meituanlongcatteam2025introducinglongcatflashthinkingtechnicalreport, meituanlongcatteam2026longcatflashthinking2601technicalreport}.

Recall the current widely used Group Relative Policy Optimization (GRPO) algorithm~\citep{grpo}, which is a variant version of Proximal Policy Optimization (PPO)~\citep{schulman2017proximalpolicyoptimizationalgorithms}.
However, applying this vanilla objective on this long-horizon task makes it face significant challenges due to \textit{distribution drift}.
Inspired by recent works~\citep{zheng2025stabilizingreinforcementlearningllms, wang2026arlarenaunifiedframeworkstable}, we can attribute this phenomenon to the train-inference discrepancy when calculating the important sampling ratio $r_{i,t}(\theta)$, which can be decomposed into two distinct manners: train-inference discrepancy $r_{i,t}^{dis}(\theta)$ and policy staleness $r_{i,t}^{stale}(\theta)$:
\begin{equation}
\begin{aligned}
r_{i,t}(\theta)=\frac{\pi_{\theta}(y_{i,t}|x,y_{i,<t})}{\mu_{\theta_{old}}(y_{i,t}|x,y_{i,<t})}=\frac{\pi_{\theta_{old}}(y_{i,t}|x,y_{i,<t})}{\mu_{\theta_{old}}(y_{i,t}|x,y_{i,<t})}\times\frac{\pi_{\theta}(y_{i,t}|x,y_{i,<t})}{\pi_{\theta_{old}}(y_{i,t}|x,y_{i,<t})}=r_{i,t}^{dis}(\theta)\times r_{i,t}^{stale}(\theta),
\end{aligned}
\end{equation}
where $\pi_{\theta}$, $\pi_{\theta_{old}}$, and $\mu_{\theta_{old}}$ denotes the current policy on train engine, old version of policy on train engine, and the old version of the policy model on the inference engine, respectively.

\begin{itemize}
    \item \textbf{Train-Inference Discrepancy} $r_{i,t}^{dis}(\theta)$. Our training infrastructure follows the asynchronous training mode that splits the whole framework into the parameter optimization (model built on Megatron engine~\citep{shoeybi2019megatron})) and experience makers (model built on vLLM engine~\citep{vllm}). 
    The difference kernels do not guarantee bitwise consistency and are especially critical when the inference and training backends are mismatched.
    In addition, {\longcatprover} is a model built on the MoE architecture. The train-inference inconsistencies may also arise due to differences in word segmentation, expert routing, and other factors.
    In general, the discrepancy between training and inference can potentially affect the calculation of the importance sampling ratio, thereby increasing the overall optimization instability risk when combined with clipping operations.
    
    \item \textbf{Policy Staleness} $r_{i,t}^{stale}(\theta)$. In asynchronous training, each generated sample may originate from multiple prior versions of the policy, which can become outdated as the current policy $\pi_\theta$ undergoes continuous updates. This discrepancy between the behavior policy that generates the data and the target policy being optimized introduces instability into the training process, hindering convergence and potentially causing model collapse in extreme cases. 
\end{itemize}

To mitigate the aforementioned issues, we introduce a hierarchical importance sampling strategy that alleviate the tokens that have significant train-inference discrepancy.
The overall surrogate objective is illustrated in the following:
\begin{equation}
\label{eq:hispo}
\begin{aligned}
\mathcal{J}_\text{GRPO}(\theta) & = \mathbb{E}_{x\sim\mathcal{D},\space y_i\sim\mu_{\theta_{old}}(\cdot|x)}\Bigg\{ 
    \textcolor{black}{\frac{1}{G\cdot\max(\{|y_i|\}_{i=1}^{G})}}
    \sum_{i=1}^G \sum_{t=1}^{|y_i|} \Big[
        \textcolor{black}{H_{i,t}(\theta)}\cdot
        \min \big(
            r_{i, t}(\theta) \hat{A}_{i, t},\textcolor{black}{\operatorname{clip}}\big(
                r_{i, t}(\theta)
            \big) \hat{A}_{i, t}
    \big)\Big] 
\Bigg\},\\
& \textcolor{black}{H_{i,t}(\theta) = \mathbb{I} \Bigg\{
    \bigg| \exp\bigg(\frac{1}{|y_i|}\sum_{j=1}^{|y_i|}\log r_{i,j}^{dis}(\theta)\bigg) - 1\bigg|< \delta_{seq}  \Bigg\} \cdot 
    \mathbb{I} \Bigg\{ \bigg |r_{i,t}^{dis}(\theta) - 1 \bigg| < \delta_{tok} \Bigg\},
}
\end{aligned}
\end{equation}
where $H_{i,t}(\theta)$ is a masking matrix that controls the gradient flow at both the sequence-level and token-level, $\mathbb{I}(\cdot)$ is the indicator function, $\delta_{seq}>0$ and $\delta_{tok}>0$ represent the hyper-parameters.

Compared to the vanilla GRPO, the major revision is to introduce a hierarchical masking strategy $H_{i,t}(\theta)$. To be specific:

\begin{itemize}
    \item \textbf{Sequence-level Discrepancy Masking}. We first estimate the sequence-level train-inference discrepancy by calculating the Geometric average of all tokens' discrepancy ratios. For the entire sequence, if its discrepancy exceeds a certain range, it is considered to have a significant impact on training stability, and the gradient contribution of the entire sequence will be removed. Compared to GSPO~\citep{gspo}, this strategy aims to consider only the sequence-level measure of training-inference consistency, avoiding the excessive neglect of valuable tokens.
    \item \textbf{Token-level Discrepancy Masking}. For the remaining sequences, we will remove tokens with significant training-pull inconsistencies to ensure that the remaining tokens will not compromise stability due to training-pull inconsistencies.
    \item \textbf{Token-level Staleness Control}. For tokens retained after sequence and token-level masking, we consider controlling staleness to ensure that the update magnitude is limited within a certain range to guarantee training stability.
\end{itemize}

We also follow our prior works~\citep{meituanlongcatteam2025introducinglongcatflashthinkingtechnicalreport, meituanlongcatteam2026longcatflashthinking2601technicalreport} to remove the divergence loss term because the use of default $k_3$ estimator, the corresponding gradient of this term is biased during optimization despite its unbiased expectation~\citep{zang2025KLGRPO}. We also employ the global constant maximum generation length during training as the denominator of the loss function. This adjustment mitigates the length bias that can pose challenges to training robustness~\citet{dr_grpo}.
Besides, as expert routing strategy may change across different versions of policies, the staleness issue can be even more obvious in sparse MoE models, where negative token-level advantages can therefore lead to excessively large importance sampling ratios and unbounded variance. 
Following~\citet{dapo} and \citet{tencent_moba}, we employ a triplet clipping scheme: $\epsilon_{\text{neg}_\text{low}}$ and $\epsilon_{\text{neg}_\text{high}}$ bound the importance ratio for negative advantages, while $\epsilon_{\text{pos}_\text{high}}$ provides an upper bound for positive advantages. This strategy prevents model collapse and maintains sufficient entropy for effective exploration.

\section{Experiments}

We conduct comprehensive evaluation of {\longcatprover} across both formal and informal reasoning tasks. 
In the formal domain, we specifically measure the model's capabilities in auto-formalization and theorem proving. 
In the informal domain, we investigate whether training on formal reasoning tasks preserves or enhances the model's performance on general reasoning tasks.
In addition, we also conduct scaling behavior analysis to show the model training performance.

\begin{table*}[t]
\setlength{\tabcolsep}{4pt}
\centering
\caption{Auto-formalization performance (Pass@8 metric, \%) of different reasoning and specific auto-formalizer models across multiple benchmarks. Best in \textbf{bold}, second in \underline{underlined}.
}
\label{tab:af_results}
\resizebox{0.88\textwidth}{!}{
\begin{tabular}{@{}l| c c c c c c c c@{}}
\toprule
\multirow{3}{*}{\textbf{Auto-Formalization}} & Combi- & FormalMath- & MathOlympiad- & MiniF2F- & ProofNet- & Prover- & Putnam- \\
 & Bench & Lite & Bench & Test & Test & Bench & Bench \\
 & $_\text{(Pass@8)}$ & $_\text{(Pass@8)}$ & $_\text{(Pass@8)}$ & $_\text{(Pass@8)}$ & $_\text{(Pass@8)}$  & $_\text{(Pass@8)}$  & $_\text{(Pass@8)}$ \\
\midrule
\rowcolor{mygray}
\multicolumn{8}{c}{\textit{Open-Weights Reasoning Models}} \\
DeepSeek-V3.2           & 65.0   & 95.2   & 85.6   & 97.5   & 81.8   & 83.0   & 46.7 \\
Kimi-K2.5               &   84.0     &   97.9     &   91.1     &   98.4     &    88.2    &    91.7    &   82.8   \\
\midrule
\rowcolor{mygray}
\multicolumn{8}{c}{\textit{Close-Weights Reasoning Models}} \\
Claude-Opus-4.5         & \underline{92.0} & 97.9   & \underline{94.4} & 98.0   & 90.9   & 94.8   & \underline{93.5} \\
Gemini-3 Pro            & 82.0   & 97.4   & 93.1   & 97.5   & \underline{91.9} & 93.0   & 90.8 \\
\midrule
\rowcolor{mygray}
\multicolumn{8}{c}{\textit{Open-Weights Auto-Formalizer Models}} \\
Kimina-Autoformalizer-7B  &  25.0 &  86.8  &  33.3  & 92.2 & 55.4 &  67.4  &  59.3\\
StepFun-Formalizer-7B  & 40.0  &   88.0 &   71.9   & 96.7 &  53.8  & 65.2 & 55.4 \\
StepFun-Formalizer-32B  &  50.0 &  90.9  &  78.6  & 95.9 & 62.4 &  73.5  & 65.1 \\
Goedel-V2-Formalizer-8B  & 61.0  & 98.1   &  73.2  &  98.4 & 76.9 & 93.0   & 80.4 \\
Goedel-V2-Formalizer-32B  & 73.0  &  98.1  & 89.2   & 98.4 & 79.0 &  94.4  & 85.9 \\
ATF-8B-Distilled  & 59.0  &   95.5 &   76.7 & 95.1  & 45.7 &   83.0 & 70.7 \\
ATF-32B  & 70.0  &  94.3  &  83.6  & 98.0 & 53.8 &  89.6  &  77.5\\
\midrule
\rowcolor{mygray}
\multicolumn{8}{c}{\textit{Ours}} \\
{\longcatprover}          & 83.0 & \underline{98.6} & 93.3 & \underline{99.2} & 87.1 & \underline{95.2} & 89.9 \\
~~w/ TIR          & \textbf{97.0} & \textbf{99.8} & \textbf{99.2} & \textbf{100.0} & \textbf{97.9} & \textbf{100.0} & \textbf{98.1} \\
\bottomrule
\end{tabular}
}
\end{table*}

\subsection{Evaluation of Auto-Formalization}
In this section, we conduct the experiments on auto-formalization tasks.
For the benchmarks, we select CombiBench~\citep{liu2025combibenchbenchmarkingllmcapability}, FormalMath-Lite~\citep{yu2025formalmathbenchmarkingformalmathematical}, MathOlympiadBench~\citep{lin2025goedelproverv2scalingformaltheorem}, MiniF2F-Test~\citep{zheng2022minif2fcrosssystembenchmarkformal}, ProofNet~\citep{azerbayev2023proofnetautoformalizingformallyproving}, ProveBench~\citep{xin2024deepseekproverv15harnessingproofassistant}, and PutnamBench~\citep{tsoukalas2024putnambenchevaluatingneuraltheoremprovers}.
The desciption of each benchmark and the corresponding inference settings are shown in Appendix~\ref{app:benchmarks}.
For the baselines, we choose two open-weights reasoning models (i.e., DeepSeek-V3.2~\citep{deepseekai2025deepseekv3technicalreport} and Kimi-K2.5~\citep{kimiteam2026kimik25visualagentic}
, two close-weights reasoning models (e.g., Claude-Opus-4.5~\citep{claude4}, and Gemini-3 Pro), and multiple open-weights auto-formalizer models (e.g., Kimina-Autoformalizer-7B~\citep{wang2025kiminaproverpreviewlargeformal}, StepFun-Formalizer-7B/32B~\citep{wu2025stepfunformalizerunlockingautoformalizationpotential}, Goedel-V2-Formalizer-8B/32B~\citep{lin2025goedelproverv2scalingformaltheorem}, ATF-8B-Distilled/32B~\citep{guo2025autoformalizertoolfeedback}).

During the evaluation process, each query is encapsulated using a uniform prompt, and the metric is Pass@8. 
We selected two verifiers that were used in the hybrid-experts iteration framework, e.g., Lean4 Server's syntax check, and LLM-as-a-Judger's semantic consistency check.
The system prompt for semantic consistency detection is shown in Appendix~\ref{app:prompts}.
The results are shown in Table~\ref{tab:af_results}. Our~{\longcatprover} establishes new state-of-the-art results on all auto-formalization benchmarks; in particular, we achieve 100\% scores on MiniF2F-Test and ProofNet. 
TIR-based enhancements yielded up to a 14\% performance gain, underscoring the efficacy of tool feedback in enabling the model to solve previously unsolvable tasks.
Furthermore, some general reasoning models demonstrated competitive performance in auto-formalization relative to proprietary auto-formalizer models. 
Notably, closed-source models such as Claude-Opus-4.5 and Gemini-3 Pro outperformed all four leading open-source models. 
Unlike proprietary auto-formalizer models, which are tailored for a single task, these closed-source models were not specifically optimized for auto-formalization, indicating that general task optimization can effectively transfer to specialized tasks.
Overall, our model outperformed both state-of-the-art general reasoning models and proprietary auto-formalizer models in auto-formalization tasks, thereby laying a robust foundation for further advancements in sketching and proving metrics.

\begin{table*}[t]
\setlength{\tabcolsep}{4pt}
\centering
\caption{Theorem-proving performance (Pass@32 metric, \%) of different reasoning and specific prover models across multiple benchmarks. Best in \textbf{bold}, second in \underline{underlined}.
$\dagger$ indicates the score is from external reports.
}
\label{tab:prove_results}
\resizebox{0.74\textwidth}{!}{
\begin{tabular}{@{}l| c c c c c c@{}}
\toprule
\multirow{3}{*}{\textbf{Theorem-Proving}} 
    & MathOlympiad- & MiniF2F- & ProofNet & Prover- & Putnam- \\
    & Bench & Test & Test &  Bench & Bench \\
    & $_\text{(Pass@32)}$ & $_\text{(Pass@32)}$ & $_\text{(Pass@32)}$  & $_\text{(Pass@32)}$  & $_\text{(Pass@32)}$ \\
\midrule
\rowcolor{mygray}
\multicolumn{6}{c}{\textit{Open-Weights Reasoning Models}} \\
DeepSeek-V3.2           & 14.7       & 77.9      & 20.4       & 42.8       & 5.8 \\
Kimi-K2.5               & 7.5       & 76.6       & 19.9       &  44.3      & 1.2     \\
\midrule
\rowcolor{mygray}
\multicolumn{6}{c}{\textit{Close-Weights Reasoning Models}} \\
Claude-Opus-4.5         &    -    &  65.6      & 35.0       &     -   &  -   \\
Gemini-3 Pro            & 3.9       & 56.2       & 22.0       & 33.5       & 1.2     \\
\midrule
\rowcolor{mygray}
\multicolumn{6}{c}{\textit{Open-Weights Prover Models}} \\
Kimina-Prover-8B  &  8.6  &  78.3$^\dagger$  &  11.8  &  37.8  &  4.8  \\
Kimina-Prover-72B  & 13.1   &  84.0$^\dagger$  &  18.3  &  44.6  &  3.9  \\
DeepSeek-Prover-V2-7B  &  11.1  &  75.6$^\dagger$  & 23.0$^\dagger$   &  49.0$^\dagger$  &  1.4$^\dagger$  \\
DeepSeek-Prover-V2-671B  &  13.9$^\dagger$  &  82.4$^\dagger$  &  30.5$^\dagger$  & 52.9$^\dagger$   & 3.3$^\dagger$   \\
Leanabell-Prover-V2-KM & -  & 68.4$^\dagger$  & 13.4$^\dagger$  & 39.8$^\dagger$  &  - \\
Leanabell-Prover-V2-DS & -  & 76.6$^\dagger$  & 23.7$^\dagger$  & 47.8$^\dagger$  &  - \\
Goedel-Prover-V2-8B  & 11.1   &  84.6$^\dagger$  & 16.7   &  49.5  & 3.8$^\dagger$  \\
~~w/ self-correction &  -  & 86.7$^\dagger$  &  -  & -   &  -  \\
Goedel-Prover-V2-32B  & 16.7$^\dagger$   &  88.1$^\dagger$  & 22.0   &  53.2  &  6.7$^\dagger$  \\
~~w/ self-correction  &  20.3$^\dagger$  &  \underline{90.4}$^\dagger$  &  -  &  -  &  8.6$^\dagger$  \\
\midrule
\rowcolor{mygray}
\multicolumn{6}{c}{\textit{Ours}} \\
{\longcatprover} \\
~~whole-proof mode  & 16.9 & 84.4 & 19.9 & 49.9 & 4.9 \\
~~whole-proof mode w/ TIR  & \underline{27.5} & 90.2  & \underline{36.1} & \underline{57.9} & \underline{10.4} \\
~~sketch-proof mode w/ TIR  & \textbf{35.8} & \textbf{93.9} & \textbf{47.3} & \textbf{66.5} & \textbf{28.9} \\
\bottomrule
\end{tabular}
}
\end{table*}

\begin{table*}[t]
\setlength{\tabcolsep}{4pt}
\centering
\caption{Theorem-proving performance (with different larger budgets, \%) of different specific prover models across multiple benchmarks. Best in \textbf{bold}, second in \underline{underlined}.
Each element $a$ / $b$ denotes to the accuracy $a$ with limited budget $b$ (i.e., Pass@$b$). ``UNK'' means unknown of the specific budget.
$\dagger$ indicates the score is from external reports.
Because different models may have different budget calculations, we directly extract the results from the report instead of conducting our own evaluations. Therefore, some benchmark results may not be available. 
}
\label{tab:prove_results_2}
\resizebox{0.98\textwidth}{!}{
\begin{tabular}{@{}l| c c c c c c@{}}
\toprule
\multirow{2}{*}{\textbf{Theorem-Proving}} 
    & MathOlympiad- & MiniF2F- & ProofNet- & Prover- & Putnam- \\
    & Bench & Test & Test &  Bench & Bench \\
\midrule
\rowcolor{mygray}
\multicolumn{6}{c}{\textit{Open-Weights Prover Models}} \\
Kimina-Prover-72B  &  - &  87.7$^\dagger$ / 1,024  & -  &  -  &  -  \\
~~w/ TTRL  &  - &  92.2$^\dagger$ / UNK  &  - &  -  &  -  \\
DeepSeek-Prover-V2-671B  & -  &  88.9$^\dagger$ / 8,192  & 37.1$^\dagger$ / 1,024  &  59.1$^\dagger$ / 512  &  7.1$^\dagger$ / 1,024  \\
Self-play Theorem Proving  &  - &  67.6$^\dagger$ / 2,560  &  26.9$^\dagger$ / 2，560 &  -  &  1.2$^\dagger$ / 3,200  \\
Leanabell-Prover-V2-KM &  - &  70.4$^\dagger$ / 128  & 18.2$^\dagger$ / 128  &  42.9$^\dagger$ / 128  &  -  \\
Leanabell-Prover-V2-DS & -  & 78.2$^\dagger$ / 128   & 25.2$^\dagger$ / 128  &  48.7$^\dagger$ / 128  &  -  \\
Goedel-Prover-V2-32B & 16.7$^\dagger$ / 32  &  92.2$^\dagger$ / 8,192  & -  &  -  &  6.7$^\dagger$ / UNK  \\
~~w/ self-correction & -  &  92.6$^\dagger$ / 1,024  & -  &  -  &  13.0$^\dagger$ / 184  \\
\midrule
\rowcolor{mygray}
\multicolumn{6}{c}{\textit{Close-Weights Prover Models}} \\
Delta-Prover &   -   &  95.9$^\dagger$ / 16,384   &    -   &   -   &  -   \\
Seed-Prover &   -   &  \textbf{99.6}$^\dagger$ / UNK   &    -   &   -   &  \underline{50.4}$^\dagger$ / UNK   \\
Seed-Prover 1.5 &   -   &  -   &    -   &   -   &  \textbf{87.9}$^\dagger$ / UNK   \\
\midrule
\rowcolor{mygray}
\multicolumn{6}{c}{\textit{Ours}} \\
{\longcatprover}  \\
~~sketch-proof mode w/ TIR  & \underline{42.5} / 180 & 95.5 / 72 & \underline{51.1} / 68 & \underline{69.5} / 220 & 31.7 / 118 \\
~~sketch-proof mode w/ TIR \& Tree Search  & \textbf{46.7} / 180 & \underline{97.1} / 72 & \textbf{52.2} / 68 & \textbf{70.8} / 220 & 41.5 / 118 \\
\bottomrule
\end{tabular}
}
\end{table*}

\subsection{Evaluation of Theorem Proving}
We further conduct the experiments on theorem proving.
For the benchmarks, we select MathOlympiadBench~\citep{lin2025goedelproverv2scalingformaltheorem}, MiniF2F-Test~\citep{zheng2022minif2fcrosssystembenchmarkformal}, ProofNet~\citep{azerbayev2023proofnetautoformalizingformallyproving}, ProveBench~\citep{xin2024deepseekproverv15harnessingproofassistant}, and PutnamBench~\citep{tsoukalas2024putnambenchevaluatingneuraltheoremprovers}, which are the competition-level open-source data for proving.
For each benchmark, we initially adopt the official formal statement as the default. 
However, we found that some of these formal statements may have semantic inconsistencies, which could prevent some proofs from being reasonably proven.
To this end, we conduct semantic consistency verification to evaluate the official formal statements. Instances of semantic inconsistency are subsequently rectified by substituting them with accurate formalizations generated by our model.

We evaluate performance across three modes. 
1) Whole-proof: we perform multiple parallel inferences per statement, reporting Pass@32 via unbiased estimation~\citep{chen2021codex}. 
Beyond Lean4 Server verification, we strictly validate theorem consistency to prevent the model from tampering with objectives or generating ``hacked'' proofs. 
2) Whole-proof with TIR: the total budget (parallel inferences $\times$ average tool calls) is capped at 32. 
3) Sketch-proof with TIR: we first sample sketches in parallel, using TIR to ensure syntactic consistency with the theorem. Each lemma within the sketch is then proven via multiple TIR inferences, with the total attempts limited to 32.

For the baselines, we choose two open-weights reasoning models (i.e., DeepSeek-V3.2~\citep{deepseekai2025deepseekv3technicalreport} and Kimi-K2.5~\citep{kimiteam2026kimik25visualagentic}, 
, two close-weights reasoning models~\footnote{Due to inherent instabilities in the inference APIs of closed-weight reasoning models, evaluation metrics may not always fully reflect their true capabilities. We are committed to providing timely updates as more stable results.} (e.g., Claude-Opus-4.5~\citep{claude4}, and Gemini-3 Pro), and multiple specific prover models (e.g., Kimina-Prover-8B/72B~\citep{wang2025kiminaproverpreviewlargeformal}, DeeSeek-Prover-V2-7B/671B~\citep{ren2025deepseekproverv2advancingformalmathematical}, Leanabell-Prover-V2-KM/DS~\citep{ji2025leanabellproverv2verifierintegratedreasoningformal}, Goedel-Prover-V2-8B/32B~\citep{lin2025goedelproverv2scalingformaltheorem}).
Since proving performance for many reasoning models was not publicly disclosed, we re-evaluated them using our own evaluation framework. For proprietary models, we adopted the official results published in their reports and supplemented these by evaluating any missing benchmarks. Throughout this process, we employed the same prompting strategy as in our previous work to maintain consistency.

We first present the performance comparison in Table~\ref{tab:prove_results} with a clear budget limit of 32 attempts (Pass@32). Our experimental results yield the following key observations:
1) In theorem-proving tasks, general-purpose reasoning models significantly underperform compared to specialized proprietary provers. This suggests that proficiency in formal theorem proving does not readily emerge from general reasoning or coding capabilities alone. 
2) {\longcatprover} demonstrates superior efficacy across different evaluation modes. In the whole-proof TIR configuration, our model surpasses the majority of baselines. In the sketch-proof TIR mode, it establishes a new state-of-the-art among all open-source provers. Specifically, on the MiniF2F-Test, we achieved a score of 93.9\%, outstripping the previous state-of-the-art, Goedel-Prover-V2, under identical computational budget constraints. Furthermore, on the more challenging PutnamBench, our model reached an accuracy of 28.9\%, significantly exceeding all other evaluated models.

We also compared the performance of tests without budget constraints. 
In this settings, we added an additional Tree Search strategy to further improve the search space of sketch-proof. For details on the implementation process, please refer to Appendix~\ref{app:tree}.
For the baselines, we externally added Self-play Theorem Proving~\citep{dong2025stpselfplayllmtheorem} (Open-weights model), Delta-Prover~\citep{zhou2025solvingformalmathproblems} (Close-weights model), Seed Prover~\citep{chen2025seedproverdeepbroadreasoning} (Close-weights model), and Seed-Prover 1.5~\citep{chen2025seedprover15masteringundergraduatelevel} (Close-weights model) for comparison. 
We only present their officially reported results to ensure the unbiasedness of the evaluation. The results are shown in Table~\ref{tab:prove_results_2}.
Our experimental results demonstrate the following:
1) {\longcatprover} outperforms all existing open-source prover models. Using MiniF2F-Test as a representative case, while Goedel-prover-v2-32B and Kimina-Prover-72B both achieved 92.2\% in whole-proof and TIR modes, they required over 1024 attempts. In contrast, our model achieved a superior score of 95.5\% with only 72 attempts, demonstrating significantly higher sample efficiency. We observe a similar performance lead across all other evaluated benchmarks.
2) Within the same budget limitation, Tree Search can further improve the accuracy, bringing an average increase of 3.1\%. This suggests that each lemma can be simplified by iteratively decomposing to make the lemma-style proof easier.
3) Comparison with proprietary models reveals competitive potential despite undisclosed constraints. 
While Seed-Prover and Seed-Prover 1.5 currently lead on MiniF2F-Test and PutnamBench, a rigorous comparison is precluded by its undisclosed and potentially much larger search budget. We intend to scale our search budget in future iterations to further narrow this gap and explore the upper bounds of our model’s performance.

\begin{table*}[t]
\setlength{\tabcolsep}{4pt}
\centering
\caption{Performance (\%) comparison across multiple general reasoning benchmarks. Best in \textbf{bold}. The result indicates that our~{\longcatprover} can retain the general reasoning ability.}
\label{tab:informal_results}
\resizebox{0.5\textwidth}{!}{
\begin{tabular}{@{}l| c c@{}}
\toprule
\multirow{3}{*}{\textbf{Informal Tasks}} 
& \textit{Prior} 
& \textit{Ours} \\
\cmidrule(lr){2-2} \cmidrule(lr){3-3}
& LongCat-Flash- 
& \textcolor{black}{LongCat-Flash-} \\
& Thinking-2601
& Prover \\
\midrule
AIME-25$_\text{(Avg@16)}$ 
& \textbf{99.6} & 97.7 \\
HMMT-25$_\text{(Avg@16)}$ 
& \textbf{93.4} & 90.8 \\
IMO-AnswerBench$_\text{(Avg@4)}$ 
& \textbf{78.6} & 77.3  \\
AMO-Bench EN$_\text{(Avg@16)}$ 
& 61.6 & \textbf{62.2} \\
AMO-Bench CH$_\text{(Avg@16)}$ 
& 56.8 & \textbf{57.3}  \\
GPQA-Diamond$_\text{(Avg@16)}$ & \textbf{80.5} & 79.2  \\
LCB (24.08-25.05)$_\text{(Avg@4)}$ & \textbf{82.8} & 81.8 \\
OJBench$_\text{(Pass@1)}$ & \textbf{42.2} & 41.8 \\
\bottomrule
\end{tabular}
}
\end{table*}

\subsection{Evaluation of General Informal Reasoning}

In this section, we aim to investigate whether our~{\longcatprover} can solve general reasoning tasks.
We select AIME-25~\citep{aime25Zhang}, HMMT-25~\citep{HMMT25}, IMO-AnswerBench~\citep{IMO-AnswerBench}, AMO-Bench with English (EN) and Chinese (CH) versions~\citep{AMO-Bench}, GPQA-Diamond~\citep{rein2024gpqa}, LiveCodeBench (24.08-25.05)~\citep{jain2025livecodebench}, and OJBench~\citep{wang2025ojbenchcompetitionlevelcode} as the benchmarks.
As illustrated in Table~\ref{tab:informal_results}, 
our method performs slightly worse on general reasoning tasks compared to~{\longcatthinknew}~\citep{meituanlongcatteam2026longcatflashthinking2601technicalreport}, suggesting that the entire training process of native formal reasoning will lead to a loss in informal reasoning. However, this loss is acceptable, and we expect to better balance the performance of native formal reasoning and informal reasoning in the later stages.

\subsection{Analysis of Reward Hacking and Evaluation Loopholes}

During the agentic RL training process of~{\longcatprover}, we observed a suspicious phenomenon that the rollout pass rate on the training set exhibited an explosive surge around the 80th step. Through careful investigation, we identified that this anomaly stemmed from critical vulnerabilities in the existing open-source evaluation pipeline. 
The existing evaluation pipeline relied solely on Lean4 syntax verification and the target theorem definition consistency checks, while the formal context of the target theorem was completely editable. This design allowed custom commands (e.g., \texttt{import}, \texttt{open}, etc.) and helper definitions (e.g., \texttt{lemma}, \texttt{instance}, etc.) to assist in the final proof of the target theorem. However, it also enabled models to cheat through various means, as summarized in Appendix~\ref{app:legality}.
These loopholes induced severe reward hacking issues and evaluation errors. To bridge the gap between the reward/metric score and the true proving capability, we developed a light-weight lexer and parser for Lean 4 proofs to convert them into ASTs, enabling rigorous inspection of cheating components. The source code of this AST-based checking is available in our project page.

\begin{figure}[htbp]
    \centering
    \begin{minipage}{0.45\textwidth}
        \centering
        \includegraphics[width=\linewidth]{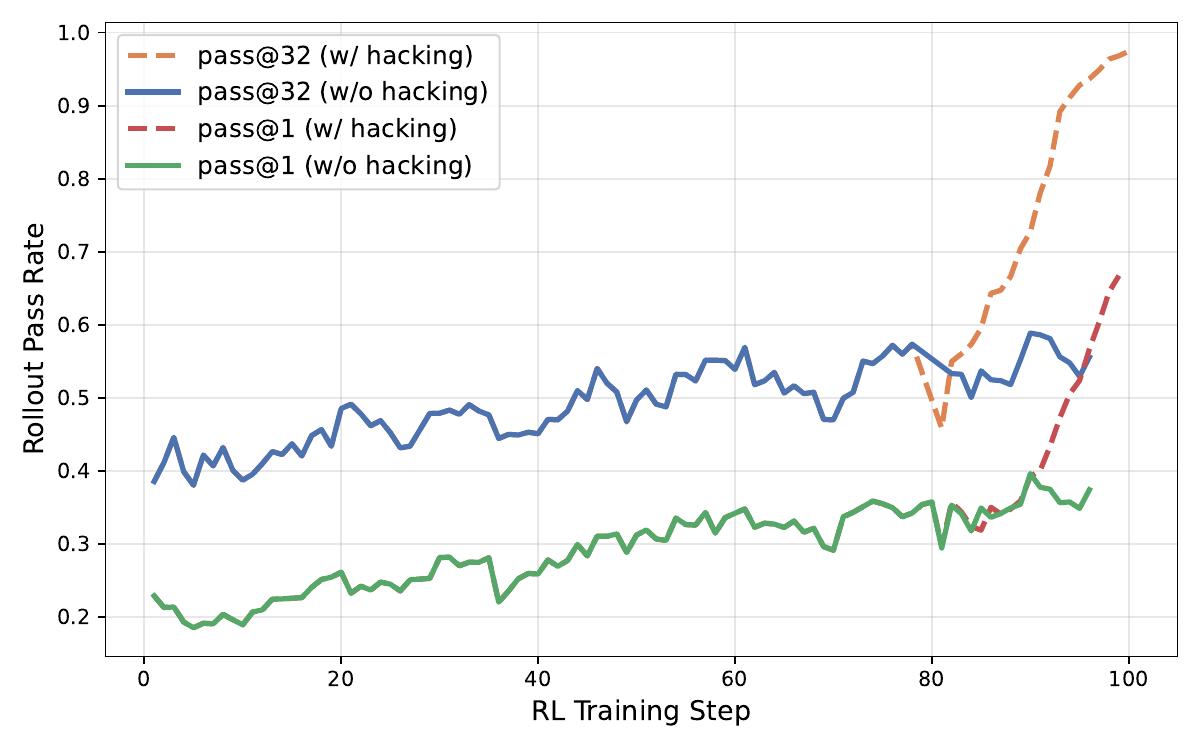}
        \caption{RL rollout pass rate (w/ v.s. w/o hacking).}
        \label{fig:rl_pass_rate}
    \end{minipage}
    \hfill
    \begin{minipage}{0.52\textwidth}
        \centering
        \captionof{table}{Evaluation across Verification Layers}
        \label{tab:verification_pass_rates}
        \scriptsize
        \begin{tabular}{lcc}
            \toprule
            \textbf{Verification Layer} & \textbf{Step-100 (hacking)} & \textbf{Step-96 (fixed)} \\
            \midrule
            Syntax Verification       & 1003 / 1024 (97.9\%)  & 715 / 1024 (69.8\%) \\
            + Target Consistency      & 999 / 1024 (97.6\%)   & 702 / 1024 (68.6\%) \\
            + AST Checking (fix) & 286 / 1024 (27.9\%)   & 499 / 1024 (48.7\%). \\
            \bottomrule
        \end{tabular}
    \end{minipage}
\end{figure}

After fixing the reward function, we resumed RL training from the 80-th step. The resulting pass rate curves on training rollouts in Figure~\ref{fig:rl_pass_rate} show that cheating behaviors were effectively suppressed, demonstrating that the reward hacking issue was mitigated with the repaired reward function.
Additionally, we generate Lean 4 proofs on 1024 training cases from both the hacking model and the fixed model, and evaluate them with different verification layers. The results are illustrated in Table~\ref{tab:verification_pass_rates}. The hacking model is shown to adeptly generate proofs that satisfy existing evaluation metrics but actually contain cheating components, whereas the fixed model is more effective at avoiding such fake proofs, ultimately achieving a higher ratio of valid proofs.

\section{Related Works}

\subsection{Large Reasoning Models (LRMs)}
In recent years, human-like reasoning like System-2 has attracted considerable attention from both academia and industry due to its significant potential for large language models (LLMs) to solve complex real-world problems through abstract and logical reasoning. 
Cutting-edge large reasoning models (LRMs), such as OpenAI's o1 \citep{openai_o1}, Google's Gemini \citep{comanici2025gemini}, DeepSeek-R1 \citep{guo2025deepseek}, and Claude code~\citep{anthropic_claude_opus_4_6}, have demonstrated the ability to engage in deep human-like thinking when solving problems by generating an extensive CoT \citep{wei2022chain} before arriving at a final solution. 
This advanced capability is typically developed through large-scale reinforcement learning with verified rewards \citep{sutton1998reinforcement, mroueh2025reinforcement, lu2025rhorizonfarlargereasoning} or self-evolutionary RL~\citep{wang2024selfevolutionarylargelanguagemodels, chen2025selfevolvingcurriculumllmreasoning, sun2025surveyneuralcodeintelligence, zhang2026memrlselfevolvingagentsruntime}. 
By scaling inference-time computation to accommodate longer chains of thought, LLMs have achieved expert-level performance on challenging tasks across mathematics, programming, and STEM disciplines. 
Recently, we released~{\longcatthink}~\citep{meituanlongcatteam2025introducinglongcatflashthinkingtechnicalreport} and~{\longcatthinknew}~\citep{meituanlongcatteam2026longcatflashthinking2601technicalreport} models. 
These models have formal reasoning capabilities, but they did not enhance the formal reasoning capabilities in the RL stage. 
Therefore, we introduce a specific model named~{\longcatprover}, which not only retains the general reasoning ability (e.g., STEM, code, agent, etc.), but also further expands the boundaries of native formal reasoning capabilities through expert iteration and agentic TIR RL.

\subsection{Formal Reasoning in LRMs}
Different from the vanilla reasoning, native formal reasoning aims to utilize the formal language (e.g., Lean4) to exhibit rationales that answer a mathematics problem, which is one of the challenging tasks in complex logic reasoning~\citep{jiangetal2025largelanguage}.
Previous work has treated auto-formalization and proving as separate models when solving formal reasoning, for example, Kimina-AutoFormalizer and Kimina-Prover-V2~\citep{wang2025kiminaproverpreviewlargeformal}, Godel-Formalizer-V2 and Goedel-Prover-V2~\citep{lin2025goedelproverv2scalingformaltheorem}, Stepfun-Formalizer~\citep{wu2025stepfunformalizerunlockingautoformalizationpotential} and Stepfun-Prover~\citep{shang2025stepfunproverpreviewletsthink}.
Early works aim to utilize the data synthesis or bootstrapping method~\citep{li2025hunyuanproverscalabledatasynthesis, wang2024theoremllamatransforminggeneralpurposellms}, tree searching strategy~\citep{han2022proofartifactcotrainingtheorem, lample2022hypertreeproofsearchneural, lamont20253dproverdiversitydriventheorem} and self-training~\citep{lin2025leanstarlearninginterleavethinking, dong2025stpselfplayllmtheorem}.
In addition, several previous efforts have been devoted to leveraging feedback from verification tools with reinforcement learning to optimize models to learn from experiences~\citep{shang2025stepfunproverpreviewletsthink, wang2025kiminaproverpreviewlargeformal, lin2025goedelproverv2scalingformaltheorem, chen2025seedproverdeepbroadreasoning, chen2025seedprover15masteringundergraduatelevel, ji2025leanabellproverv2verifierintegratedreasoningformal, shen2025realproverretrievalaugmentedlean, xin2024deepseekproverv15harnessingproofassistant, ren2025deepseekproverv2advancingformalmathematical, xin2025scalingmultiturnoffpolicyrl}.
In contrast to these approaches, we treat formal reasoning and informal reasoning as an integrated learning mode, and consider auto-formalization, sketching, and proving as atomic capabilities of native formal reasoning. These capabilities can be composably applied to solve complex problems. Furthermore, we enhance the upper bound of formal reasoning performance through large-scale TIR RL.

\section{Conclusion}

In this work, we introduce~{\longcatprover}, a $560$-billion-parameter Mixture-of-Experts (MoE) model that fuses the native formal reasoning with general reasoning capabilities.
It achieves state-of-the-art performance among open-source models on multiple auto-formalization and theorem-proving tasks with verified tools. 
The core innovations underpinning {\longcatprover} are as follows:
1) an effective hybrid-experts iteration framework to better construct high-quality synthesized trajectories with multiple verified tools.
2) a Hierarchical Importance Sampling Policy Optimization method that stabilizes the RL training based on MoE model.
We also explore multiple verified tools for formal reasoning and futher ensure the reliability of the theorem proving by legality detection.
We hope that the open-sourcing of~{\longcatprover} will advance research in both informal reasoning and formal reasoning, particularly in the areas of high-quality data strategies, efficient RL training, and native agentic reasoning.

\newpage

\begin{CJK}{UTF8}{gbsn}

\section{Contributions}

\subsection*{Core Contributors}

\emph{An asterisk (*) indicates members who have equal contributions.}

Jianing Wang *, Jianfei Zhang *, Qi Guo *, Linsen Guo, Rumei Li

\subsection*{Tech Leads}
Wei Wang, Peng Pei, Xunliang Cai

\subsection*{Contributions}

\emph{We are grateful for all the support from the following partners, listed in alphabetical order by first name. A dagger ($\dagger$) indicates members who have left the LongCat team.}

Chao Zhang, Chong Peng, Cunguang Wang, Dengchang Zhao, Jiarong Shi, Jingang Wang, Liulin Feng, Mengxia Shen, Qi Li, Shengnan An, Shun Wang$\dagger$, Wei Shi, Xiangyu Xi, Xiaoyu Li, Xuezhi Cao, Yi Lu, Yunke Zhao, Zhengyu Chen, Zhimin Lin

\end{CJK}

\newpage

\bibliographystyle{unsrtnat}
\bibliography{references}

\newpage
\appendix
\def\lstlanguagefiles{./utils/lstlean.tex}

\lstset{
    upquote=true,
    inputencoding=utf8,
    extendedchars=true,
    basicstyle=\footnotesize\ttfamily,
    frame=none,
    breaklines=true,
    keepspaces=true,
    mathescape=false,
    columns=flexible,
    breakindent=0pt,
    literate={∀}{{\ensuremath{\forall}}}1
             {∃}{{\ensuremath{\exists}}}1
             {∩}{{\ensuremath{\cap}}}1
             {∪}{{\ensuremath{\cup}}}1
             {∅}{{\ensuremath{\emptyset}}}1
             {∧}{{\ensuremath{\land}}}1
             {⊢}{{\ensuremath{\vdash}}}1
             {∨}{{\ensuremath{\lor}}}1
             {→}{{\ensuremath{\rightarrow}}}1
             {←}{{\ensuremath{\leftarrow}}}1
             {↔}{{\ensuremath{\leftrightarrow}}}1
             {∈}{{\ensuremath{\in}}}1
             {∉}{{\ensuremath{\notin}}}1
             {≤}{{\ensuremath{\leq}}}1
             {≥}{{\ensuremath{\geq}}}1
             {≠}{{\ensuremath{\neq}}}1
             {≡}{{\ensuremath{\equiv}}}1
             {⊆}{{\ensuremath{\subseteq}}}1
             {⊇}{{\ensuremath{\supseteq}}}1
             {⊂}{{\ensuremath{\subset}}}1
             {⊃}{{\ensuremath{\supset}}}1
             {¬}{{\ensuremath{\neg}}}1
             {∑}{{\ensuremath{\sum}}}1
             {∏}{{\ensuremath{\prod}}}1
             {∫}{{\ensuremath{\int}}}1
             {√}{{\ensuremath{\sqrt{}}}}1
             {ℕ}{{\ensuremath{\mathbb{N}}}}1
             {ℤ}{{\ensuremath{\mathbb{Z}}}}1
             {ℝ}{{\ensuremath{\mathbb{R}}}}1
             {ℚ}{{\ensuremath{\mathbb{Q}}}}1
             {α}{{\ensuremath{\alpha}}}1
             {β}{{\ensuremath{\beta}}}1
             {γ}{{\ensuremath{\gamma}}}1
             {λ}{{\ensuremath{\lambda}}}1
             {'}{{'}}1
             {'}{{\textquoteleft}}1
             {"}{{\textquotedblleft}}1
             {"}{{\textquotedblright}}1
             {…}{{\ensuremath{\ldots}}}1
             {·}{{\ensuremath{\cdot}}}1
             {₀}{{\textsubscript{0}}}1
             {₁}{{\textsubscript{1}}}1
             {₂}{{\textsubscript{2}}}1
             {₃}{{\textsubscript{3}}}1
             {₄}{{\textsubscript{4}}}1
             {₅}{{\textsubscript{5}}}1
             {₆}{{\textsubscript{6}}}1
             {₇}{{\textsubscript{7}}}1
             {₈}{{\textsubscript{8}}}1
             {₉}{{\textsubscript{9}}}1
             {ᵢ}{{\textsubscript{i}}}1
             {—}{{---}}1
             {⟨}{{\ensuremath{\langle}}}1
             {⟩}{{\ensuremath{\rangle}}}1
             {↑}{{\ensuremath{\uparrow}}}1
             {↓}{{\ensuremath{\downarrow}}}1
             {✝}{{\ensuremath{\dagger}}}1
             {×}{{\ensuremath{\times}}}1
             {÷}{{\ensuremath{\div}}}1
}

\section{Details of Benchmarks}
\label{app:benchmarks}

To comprehensively evaluate our method, we select a diverse collection of mathematical benchmarks. All of these benchmarks contain natural language problem statements paired with their formal counterparts an are selected for \textbf{auto-formalization} evaluation. For \textbf{theorem-proving}, we specifically select MathOlympiad, MiniF2F, ProofNet, ProverBench, and PutnamBench, as these datasets are extensively utilized by existing prover models.

\begin{itemize}\setlength{\itemsep}{0pt}
    \item \textbf{Combibench}: Combibench comprises 100 combinatorial math problems, covering a wide spectrum of difficulty levels, ranging from middle school to university and International Mathematical Olympiad (IMO) levels, and spans over ten combinatorial topics. License: MIT.
    \item \textbf{FormalMath-lite}: A carefully selected subset of 425 problems (359 high school and 66 undergraduate level) derived from the rigorously validated FormalMATH benchmark (5,560 problems), where each problem is verified through a hybrid pipeline of multi-LLM semantic verification and careful review by Olympiad-level experts. License: MIT.
    \item \textbf{MathOlympiad-Bench}: A high-quality dataset comprising 360 human-verified formalizations of Olympiad-level mathematical competition problems, including 158 IMO problems from 1959 to 2024, 131 IMO shortlist problems covering 2006 to 2023, 68 national mathematical Olympiad problems, and 3 additional mathematical puzzles. License: Apache-2.0.
    \item \textbf{MiniF2F}: A widely recognized dataset created for evaluating automated theorem proving techniques, featuring a collection of mathematical problems designed for human challenge. It consists of 244 training problems and 244 test problems. In our experiments, we perform the evaluation exclusively on the test split. License: Apache-2.0.
    \item \textbf{ProofNet}: ProofNet consists of 371 problems in Lean 3, spanning popular undergraduate pure mathematics textbooks and topics such as real and complex analysis, linear algebra, abstract algebra, and topology. We select the Lean 4 translation of ProofNet made available by~\citep{xin2024deepseekproverv15harnessingproofassistant}, which is further divided into two splits: ProofNet-valid and ProofNet-test, containing 185 and 186 problems, respectively. We choose the test split of ProofNet for model evaluation. License: MIT.
    \item \textbf{ProverBench}: ProverBench contains 325 formalized mathematical problems designed to enable more comprehensive evaluation across both high-school competition problems and undergraduate-level mathematics. The collection features 15 authentic high-school competition challenges sourced from recent AIME competitions in number theory and algebra, alongside 310 pedagogically grounded problems drawn from curated textbook examples and educational tutorials. License: MIT.
    \item \textbf{PutnamBench}: A dataset derived from the prestigious Putnam Mathematical Competition, featuring problems sourced from 1965 to 2025. It comprises 672 formalizations in Lean 4. License: MIT.
\end{itemize}



\section{Task Cases}
\label{app:cases}

\subsection{Case of Auto-formalization without Tools}

\begin{tcolorbox}[
    colback=white,      
    colframe=black,     
    arc=5pt,            
    boxrule=1pt,        
    width=\linewidth,   
    center,             
    left=0mm,           
    right=0mm,          
    top=0mm,            
    bottom=0mm,         
    title={Case: Ladies' Club Partition into Two Cliques},
    breakable
]

\textbf{Informal Statement:}

There are 100 members in a ladies' club. Each lady has had tea (in private) with exactly 56 of the other members of the club. The Board, consisting of the 50 most distinguished ladies, has all had tea with one another. Prove that the entire club may be split into two groups in such a way that, within each group, any lady has had tea with any other.

\textbf{Generation without Tools:}
\begin{lstlisting}
<longcat_think>
Let me understand this problem:

1. There are 100 members in a ladies' club
2. Each lady has had tea with exactly 56 of the other members
3. The Board consists of the 50 most distinguished ladies
4. The Board members have all had tea with one another (so the Board forms a complete graph)
5. We need to prove that the entire club can be split into two groups such that within each group, any lady has had tea with any other (i.e., each group forms a complete graph)

Let me think about how to formalize this in Lean 4.

First, I need to represent:
- A set of 100 ladies
- A tea-drinking relation (a graph)
- The constraint that each lady has had tea with exactly 56 others
- The constraint that the Board (50 ladies) forms a complete graph
- The conclusion that the club can be partitioned into two complete subgraphs

Let me structure this:

```lean4
theorem my_theorem : 
  ∀ (ladies : Finset ℕ) (tea : ℕ → Finset ℕ),
    ladies.card = 100 →
    (∀ l ∈ ladies, tea l ⊆ ladies) →
    (∀ l ∈ ladies, (tea l).card = 56) →
    (∃ board : Finset ℕ, board ⊆ ladies ∧ board.card = 50 ∧ 
      ∀ a b, a ∈ board → b ∈ board → a ≠ b → b ∈ tea a) →
    ∃ (A B : Finset ℕ), A ∪ B = ladies ∧ A ∩ B = ∅ ∧
      (∀ a b, a ∈ A → b ∈ A → a ≠ b → b ∈ tea a) ∧
      (∀ a b, a ∈ B → b ∈ B → a ≠ b → b ∈ tea a) := by sorry
```

I'm realizing the complexity of the current formalization might be overwhelming. The core challenge is finding a way to partition the ladies into two groups where each group forms a complete graph of tea-drinking relationships.

The key insights are:
- We have 100 total ladies
- Each lady drinks tea with exactly 56 others
- A board of 50 ladies has already established complete tea-drinking connections
- We need to prove we can split the remaining ladies into two complete tea-drinking groups

The problem requires a clever partitioning strategy that leverages the existing tea-drinking relationships to create two fully connected subgroups.

I'll need to construct a proof that demonstrates how to divide the ladies into two sets where every member within each set has tea-drunk with every other member, using the initial board's complete connectivity as a strategic starting point.

The challenge lies in showing that the remaining 50 ladies can be perfectly divided into two 50-member groups with complete internal tea-drinking connections, while maintaining the original board's complete graph structure.

The key constraints are:
- Total ladies: 100
- Each lady drinks tea with exactly 56 others
- Board of 50 ladies has complete internal tea-drinking
- Goal is to partition the entire set into two complete subgraphs

The proof strategy will involve carefully analyzing the tea-drinking relationships and leveraging the board's complete graph to construct the desired partition.

I'll need to prove that the remaining 50 ladies can be split into two groups where each group has complete internal tea-drinking connections. The challenge is ensuring each group maintains the 56-tea-drunk constraint while being fully interconnected.

The theorem requires showing the existence of two disjoint sets A and B that together cover all 100 ladies, with each set having complete internal tea-drinking relationships.

The key constraints are:
- Total ladies: 100
- Each lady drinks tea with exactly 56 others
- A board of 50 ladies has complete tea-drinking connections
- Goal is to partition the remaining 50 ladies into two complete tea-drinking groups

The proof strategy involves demonstrating how the initial board's complete graph can be used to construct the desired partition, ensuring each subset maintains the required tea-drinking properties.

The theorem seeks to prove the existence of two disjoint sets A and B that together cover the remaining 50 ladies, with each set having complete internal tea-drinking relationships.

The key constraints involve the tea-drinking connections between ladies, specifically that each lady drinks tea with exactly 56 others, and that the board forms a complete graph. The goal is to partition the remaining 50 ladies into two groups where each group maintains complete internal tea-drinking connectivity.

The proof strategy involves demonstrating that such a partition is possible under these specific conditions of tea-drinking relationships and set composition.
</longcat_think>
Looking at this problem, I need to formalize:
1. A club of 100 ladies with a tea-drinking relation
2. Each lady drinks tea with exactly 56 others  
3. The Board of 50 ladies forms a complete graph
4. The conclusion that the club can be partitioned into two complete subgraphs

```lean4
import Mathlib
import Aesop

set_option maxHeartbeats 0

open BigOperators Real Nat Topology Rat

theorem my_theorem : 
  ∀ (tea : Fin 100 → Finset (Fin 100)),
    (∀ i : Fin 100, (tea i).card = 56) →
    (∃ board : Finset (Fin 100), board.card = 50 ∧ 
      ∀ a b, a ∈ board → b ∈ board → a ≠ b → b ∈ tea a) →
    ∃ (A B : Finset (Fin 100)), A ∪ B = Finset.univ ∧ A ∩ B = ∅ ∧
      (∀ a b, a ∈ A → b ∈ A → a ≠ b → b ∈ tea a) ∧
      (∀ a b, a ∈ B → b ∈ B → a ≠ b → b ∈ tea a) := by sorry
```

\end{lstlisting}

\end{tcolorbox}

\subsection{Case of Auto-formalization with Tools}

\begin{tcolorbox}[
    colback=white,      
    colframe=black,     
    arc=5pt,            
    boxrule=1pt,        
    width=\linewidth,   
    center,             
    left=0mm,           
    right=0mm,          
    top=0mm,            
    bottom=0mm,         
    title={Case: Father Give the Balls},
    breakable
]

\textbf{Informal Statement:}

The father has six sons and ten identical, indistinguishable balls. How many ways can he give the balls to his sons if everyone gets at least one? Prove that the answer is 126.

\textbf{Generation with Tools:}
\begin{lstlisting}
<longcat_think>
I need to analyze this combinatorial problem and convert it to Lean 4.

This is a classic "stars and bars" problem with constraints. We need to distribute 10 identical balls among 6 sons, where each son gets at least one ball.

The mathematical approach:
- Since each son must get at least one ball, we first give each son 1 ball (6 balls total)
- We have 4 balls remaining to distribute freely among 6 sons
- This becomes equivalent to finding non-negative integer solutions to x₁ + x₂ + x₃ + x₄ + x₅ + x₆ = 4
- The number of such solutions is C(4+6-1, 6-1) = C(9,5) = 126

Let me write the Lean 4 statement:

```lean4
import Mathlib
import Aesop

theorem my_favorite_theorem : 
  (Finset.filter (fun f : Fin 6 → Fin 10 => 
    (∀ i : Fin 6, 1 ≤ (Finset.univ.filter (fun j => f j = i)).card) ∧ 
    (Finset.univ.image f).card = 10) 
    (Finset.univ : Finset (Fin 6 → Fin 10))).card = 126 := by sorry
```
</longcat_think>
<longcat_tool_call>syntax_check
<longcat_arg_key>formal_statement</longcat_arg_key>
<longcat_arg_value>import Mathlib
import Aesop

theorem my_favorite_theorem : 
  (Finset.filter (fun f : Fin 6 → Fin 10 => 
    (∀ i : Fin 6, 1 ≤ (Finset.univ.filter (fun j => f j = i)).card) ∧ 
    (Finset.univ.image f).card = 10) 
    (Finset.univ : Finset (Fin 6 → Fin 10))).card = 126 := by sorry</longcat_arg_value>
</longcat_tool_call>
</longcat_s><longcat_observation><longcat_tool_response>syntax_check
{"pass": true, "errors": []}</longcat_tool_response> /think_on <longcat_assistant><longcat_think>
Great! The syntax check passed. Now I need to verify that this statement is consistent with the original problem.
</longcat_think>
<longcat_tool_call>consistency_check
<longcat_arg_key>formal_statement</longcat_arg_key>
<longcat_arg_value>import Mathlib
import Aesop

theorem my_favorite_theorem : 
  (Finset.filter (fun f : Fin 6 → Fin 10 => 
    (∀ i : Fin 6, 1 ≤ (Finset.univ.filter (fun j => f j = i)).card) ∧ 
    (Finset.univ.image f).card = 10) 
    (Finset.univ : Finset (Fin 6 → Fin 10))).card = 126 := by sorry</longcat_arg_value>
<longcat_arg_key>informal_statement</longcat_arg_key>
<longcat_arg_value>The father has six sons and ten identical, indistinguishable balls. How many ways can he give the balls to his sons if everyone gets at least one?
Prove that the answer is 126</longcat_arg_value>
</longcat_tool_call>
</longcat_s><longcat_observation><longcat_tool_response>consistency_check
{"pass": false, "explanations": "1. Mathematical Text Analysis: The problem asks for the number of ways to distribute 10 indistinct balls to 6 distinguishable sons with each son getting at least one. This is a classic stars-and-bars problem (variables: non-negative integers x₁,...,x₆ with ∑x_i = 10 and x_i ≥ 1). The expected solution uses the formula C(n-1,k-1) = C(9,5) = 126.\n\n2. Lean4 Code Analysis: The theorem formalizes functions f : Fin 6 → Fin 10 (sons to balls?), filtered by two conditions: (1) every son (Fin 6) receives at least one ball (count of j where f j = i ≥1), and (2) the image of f has cardinality exactly 10. The Finset.univ here represents all functions from Fin 6 to Fin 10.\n\n3. Comparative Analysis: \n- The problem requires distributing *indistinct* balls, but the Lean code models balls as distinguishable (since it tracks individual assignments via Fin 10 → Fin 6 functions).\n- The second condition (image cardinality = 10) is impossible since there are only 6 sons (Fin 6), making this a logical inconsistency.\n- The correct constraint for the problem would require the sum of all assigned balls (not image cardinality) to equal 10, and each son's count ≥1.\n- The function type Fin 6 → Fin 10 is semantically mismatched: the codomain Fin 10 does not represent ball counts but seems to incorrectly model ball indices.\n\n4. Accuracy Confirmation: The formalization is incorrect due to fundamental mismatches in problem modeling, logical constraints, and semantic misinterpretation of variables."}</longcat_tool_response> /think_on <longcat_assistant><longcat_think>
The consistency check failed because my formalization incorrectly modeled the problem. I need to fix the semantic issues:

1. I'm modeling balls as distinguishable rather than identical
2. The function type is backwards and doesn't represent the actual distribution
3. The constraints don't properly represent the problem requirements

Let me fix this by using the correct combinatorial approach where we find the number of non-negative integer solutions to x₁ + x₂ + ... + x₆ = 10 with each xᵢ ≥ 1.

```lean4
import Mathlib
import Aesop

theorem my_favorite_theorem : 
  (Finset.filter (fun v : Fin 6 → ℕ => 
    (∀ i : Fin 6, v i ≥ 1) ∧ 
    (Finset.univ.sum v) = 10) 
    (Finset.univ : Finset (Fin 6 → ℕ))).card = 126 := by sorry
```
</longcat_think><longcat_tool_call>syntax_check
<longcat_arg_key>formal_statement</longcat_arg_key>
<longcat_arg_value>import Mathlib
import Aesop

theorem my_favorite_theorem : 
  (Finset.filter (fun v : Fin 6 → ℕ => 
    (∀ i : Fin 6, v i ≥ 1) ∧ 
    (Finset.univ.sum v) = 10) 
    (Finset.univ : Finset (Fin 6 → ℕ))).card = 126 := by sorry</longcat_arg_value>
</longcat_tool_call>
</longcat_s><longcat_observation><longcat_tool_response>syntax_check
{"pass": false, "errors": [{"severity": "error", "pos": {"line": 5, "column": 5}, "endPos": {"line": 5, "column": 16}, "data": "failed to synthesize\n  Fintype (Fin 6 → ℕ)\nAdditional diagnostic information may be available using the `set_option diagnostics true` command."}]}</longcat_tool_response> /think_on <longcat_assistant><longcat_think>
The syntax check failed because `Finset.univ : Finset (Fin 6 → ℕ)` doesn't work since functions from `Fin 6` to `ℕ` don't form a finite set. I need to use a different approach that properly represents the finite combinatorial problem.

Let me use a different formulation that explicitly represents the stars-and-bars constraint:

```lean4
import Mathlib
import Aesop

theorem my_favorite_theorem : 
  (Finset.filter (fun v : Fin 6 → Fin 11 => 
    (∀ i : Fin 6, v i ≥ 1) ∧ 
    (Finset.univ.sum (fun i => (v i : ℕ))) = 10) 
    (Finset.univ : Finset (Fin 6 → Fin 11))).card = 126 := by sorry
```
</longcat_think><longcat_tool_call>syntax_check
<longcat_arg_key>formal_statement</longcat_arg_key>
<longcat_arg_value>import Mathlib
import Aesop

theorem my_favorite_theorem : 
  (Finset.filter (fun v : Fin 6 → Fin 11 => 
    (∀ i : Fin 6, v i ≥ 1) ∧ 
    (Finset.univ.sum (fun i => (v i : ℕ))) = 10) 
    (Finset.univ : Finset (Fin 6 → Fin 11))).card = 126 := by sorry</longcat_arg_value>
</longcat_tool_call>
</longcat_s><longcat_observation><longcat_tool_response>syntax_check
{"pass": true, "errors": []}</longcat_tool_response> /think_on <longcat_assistant><longcat_think>
Great! The syntax check passed. Now I need to verify that this statement is consistent with the original problem.
</longcat_think>
<longcat_tool_call>consistency_check
<longcat_arg_key>formal_statement</longcat_arg_key>
<longcat_arg_value>import Mathlib
import Aesop

theorem my_favorite_theorem : 
  (Finset.filter (fun v : Fin 6 → Fin 11 => 
    (∀ i : Fin 6, v i ≥ 1) ∧ 
    (Finset.univ.sum (fun i => (v i : ℕ))) = 10) 
    (Finset.univ : Finset (Fin 6 → Fin 11))).card = 126 := by sorry</longcat_arg_value>
<longcat_arg_key>informal_statement</longcat_arg_key>
<longcat_arg_value>The father has six sons and ten identical, indistinguishable balls. How many ways can he give the balls to his sons if everyone gets at least one?
Prove that the answer is 126</longcat_arg_value>
</longcat_tool_call>
</longcat_s><longcat_observation><longcat_tool_response>consistency_check
{"pass": true, "explanations": "1. Mathematical Text Analysis: The problem involves distributing 10 identical balls to 6 sons, with each son receiving at least one ball. The answer is 126, derived from the stars-and-bars combinatorial formula C(10-1,6-1) = C(9,5) = 126. Key elements: 6 sons, 10 balls, identical items, at least one per son, and the count of distributions.\n\n2. Lean4 Code Analysis: The theorem defines a set of functions from Fin 6 to Fin 11, filtered by two conditions: (1) all outputs are ≥1, and (2) the sum of outputs equals 10. It claims the cardinality of this set is 126. The codomain Fin 11 allows values 0 to 10, but the filter enforces outputs ≥1. The sum constraint ensures the total is 10.\n\n3. Comparative Analysis: The formalization aligns with the problem. The 6 sons correspond to Fin 6, the balls to the sum of function outputs (10), and the constraint that each son gets at least one ball is enforced by ∀ i, v i ≥ 1. The codomain Fin 11 is sufficient (though slightly overinclusive) because the filter excludes invalid values. The count of valid functions (126) matches the answer. No constraints or variables are missing or misaligned. 4. Accuracy Confirmation: All elements match. 
The Lean code correctly models the combinatorial structure, including the number of sons, the total balls, the minimum per son, and the expected count. The use of Fin 11 as the codomain does not affect correctness because the filter ensures only valid distributions are counted."}</longcat_tool_response> /think_on <longcat_assistant><longcat_think>
Excellent! Both verification checks have passed successfully. The statement correctly formalizes the combinatorial problem of distributing 10 identical balls among 6 sons where each son gets at least one ball.

The final Lean 4 statement is:

```lean4
import Mathlib
import Aesop

theorem my_favorite_theorem : 
  (Finset.filter (fun v : Fin 6 → Fin 11 => 
    (∀ i : Fin 6, v i ≥ 1) ∧ 
    (Finset.univ.sum (fun i => (v i : ℕ))) = 10) 
    (Finset.univ : Finset (Fin 6 → Fin 11))).card = 126 := by sorry
```

This statement successfully captures the mathematical structure of the problem:
- `Fin 6` represents the 6 sons
- Functions from `Fin 6 → Fin 11` represent possible distributions (allowing values 0-10 for each son)
- The filter ensures each son gets at least one ball (`∀ i, v i ≥ 1`)
- The sum constraint ensures exactly 10 balls are distributed
- The result is that there are exactly 126 such valid distributions


</longcat_think>
***Final Lean4 Formalization***
```lean4
import Mathlib
import Aesop

theorem my_favorite_theorem : 
  (Finset.filter (fun v : Fin 6 → Fin 11 => 
    (∀ i : Fin 6, v i ≥ 1) ∧ 
    (Finset.univ.sum (fun i => (v i : ℕ))) = 10) 
    (Finset.univ : Finset (Fin 6 → Fin 11))).card = 126 := by sorry
```

\end{lstlisting}

\end{tcolorbox}

\subsection{Case of Theorem Proving}
\begin{tcolorbox}[
    colback=white,      
    colframe=black,     
    arc=5pt,            
    boxrule=1pt,        
    width=\linewidth,   
    center,             
    left=0mm,           
    right=0mm,          
    top=0mm,            
    bottom=0mm,         
    title={Case: Putnam\_1990\_a1},
    breakable
]

\textbf{Informal Statement:}
Let $T_0=2,T_1=3,T_2=6$, and for $n \geq 3$, $T_n=(n+4)T_{n-1}-4nT_{n-2}+(4n-8)T_{n-3}$. The first few terms are $2,3,6,14,40,152,784,5168,40576$. Find, with proof, a formula for $T_n$ of the form $T_n=A_n+B_n$, where $\{A_n\}$ and $\{B_n\}$ are well-known sequences.

\textbf{Formal Statement:}
\begin{lstlisting}[language=lean]
import Mathlib

open Filter Topology Nat

abbrev putnam_1990_a1_solution : (ℕ → ℤ) × (ℕ → ℤ) := (fun n : ℕ => (n)!, fun n : ℕ => 2 ^ n)

theorem putnam_1990_a1
    (T : ℕ → ℤ)
    (hT012 : T 0 = 2 ∧ T 1 = 3 ∧ T 2 = 6)
    (hTn : ∀ n, T (n + 3) = (n + 7) * T (n + 2) - 4 * (n + 3) * T (n + 1) + (4 * n + 4) * T n) :
    T = putnam_1990_a1_solution.1 + putnam_1990_a1_solution.2 :=
  sorry
\end{lstlisting}

\textbf{Generation:}

\begin{lstlisting}
<longcat_think>
# Solving the Sequence Problem in Lean 4

We need to find a formula for $T_n$ where $T_0=2$, $T_1=3$, $T_2=6$, and for $n \geq 3$, $T_n=(n+4)T_{n-1}-4nT_{n-2}+(4n-8)T_{n-3}$. The final answer should be $T_n = n! + 2^n$.

First, let's verify the formula $T_n = n! + 2^n$ for the initial terms:

- For $n=0$: $0! + 2^0 = 1 + 1 = 2$ which matches $T_0 = 2$.
- For $n=1$: $1! + 2^1 = 1 + 2 = 3$ which matches $T_1 = 3$.
- For $n=2$: $2! + 2^2 = 2 + 4 = 6$ which matches $T_2 = 6$.
- For $n=3$: $3! + 2^3 = 6 + 8 = 14$ which matches the given $T_3 = 14$.
- For $n=4$: $4! + 2^4 = 24 + 16 = 40$ which matches $T_4 = 40$.
- For $n=5$: $5! + 2^5 = 120 + 32 = 152$ which matches $T_5 = 152$.

Now let's check if the recurrence relation holds for $T_n = n! + 2^n$:

For $n \geq 3$, we need to show:
$T_n = (n+4)T_{n-1} - 4nT_{n-2} + (4n-8)T_{n-3}$

Substituting $T_k = k! + 2^k$:
- Left side: $n! + 2^n$
- Right side: $(n+4)((n-1)! + 2^{n-1}) - 4n((n-2)! + 2^{n-2}) + (4n-8)((n-3)! + 2^{n-3})$

Let's simplify the right side:

First, expand each term:
1. $(n+4)( (n-1)! + 2^{n-1} ) = (n+4)(n-1)! + (n+4)2^{n-1}$
2. $-4n( (n-2)! + 2^{n-2} ) = -4n(n-2)! -4n2^{n-2}$
3. $+(4n-8)( (n-3)! + 2^{n-3} ) = (4n-8)(n-3)! + (4n-8)2^{n-3}$

Now let's simplify term by term.

First, the factorial terms:
- $(n+4)(n-1)! = (n+4)(n-1)(n-2)(n-3)!$
- $-4n(n-2)! = -4n(n-2)(n-3)!$
- $+(4n-8)(n-3)!$

So combining these:
$(n+4)(n-1)(n-2)(n-3)! -4n(n-2)(n-3)! + (4n-8)(n-3)!$

Factor $(n-3)!$:
$(n-3)! [ (n+4)(n-1)(n-2) -4n(n-2) + (4n-8) ]$

Expand the polynomial:
$(n+4)(n-1)(n-2) = (n+4)(n^2 - 3n + 2) = n^3 - 3n^2 + 2n + 4n^2 - 12n + 8 = n^3 + n^2 - 10n + 8$

Then:
$-4n(n-2) = -4n^2 + 8n$

Then:
$+ (4n - 8)$

So total:
$(n^3 + n^2 - 10n + 8) + (-4n^2 + 8n) + (4n - 8) = n^3 - 3n^2 + 2n$

So the factorial terms give:
$(n-3)! (n^3 - 3n^2 + 2n) = (n-3)! \cdot n(n^2 - 3n + 2) = (n-3)! \cdot n(n-1)(n-2) = n!$

So that's good, the factorial terms give $n!$.

Now the exponential terms:
$(n+4)2^{n-1} -4n2^{n-2} + (4n-8)2^{n-3}$

Factor $2^{n-3}$:
$2^{n-3} [ (n+4)2^{2} -4n2^{1} + (4n-8) ]$

Compute the expression in the brackets:
$(n+4)4 = 4n + 16$
$-4n \cdot 2 = -8n$
$+ (4n - 8)$

So total:
$4n + 16 -8n + 4n - 8 = (4n -8n +4n) + (16 -8) = 0n + 8 = 8$

So the exponential terms give $2^{n-3} \cdot 8 = 2^{n-3} \cdot 2^3 = 2^n$.

So total right side is $n! + 2^n$ which equals left side.

Therefore the formula holds.

Now let's formalize this in Lean 4.

```tactics
import Mathlib

open Filter Topology Nat

abbrev putnam_1990_a1_solution : (ℕ → ℤ) × (ℕ → ℤ) := (fun n : ℕ => (n)!, fun n : ℕ => 2 ^ n)
```

So we define the pair of functions as $n \mapsto n!$ and $n \mapsto 2^n$.

```tactics
/--
Let $T_0=2,T_1=3,T_2=6$, and for $n \geq 3$, $T_n=(n+4)T_{n-1}-4nT_{n-2}+(4n-8)T_{n-3}$. The first few terms are $2,3,6,14,40,152,784,5168,40576$. Find, with proof, a formula for $T_n$ of the form $T_n=A_n+B_n$, where $\{A_n\}$ and $\{B_n\}$ are well-known sequences.
-/
theorem putnam_1990_a1
    (T : ℕ → ℤ)
    (hT012 : T 0 = 2 ∧ T 1 = 3 ∧ T 2 = 6)
    (hTn : ∀ n, T (n + 3) = (n + 7) * T (n + 2) - 4 * (n + 3) * T (n + 1) + (4 * n + 4) * T n) :
    T = putnam_1990_a1_solution.1 + putnam_1990_a1_solution.2 := by
```

Now we need to prove that $T_n = n! + 2^n$ for all $n$.

```tactics
have h1 : ∀ (n : ℕ), T n = (n : ℤ) ! + (2 : ℤ) ^ n := by
```

We will prove this by induction.

```tactics
intro n
induction' n using Nat.strongRecOn with n ih
```

Now we need to consider the base cases and the induction step.

```tactics
cases n with
| zero =>
```

First base case $n = 0$.

We need to show $T 0 = 0! + 2^0 = 1 + 1 = 2$.

```tactics
rcases hT012 with ⟨h0, h1, h2⟩
simp [h0]
```

So it works.

```tactics
| succ n =>
  cases n with
  | zero =>
```

Second base case $n = 1$.

We need to show $T 1 = 1! + 2^1 = 1 + 2 = 3$.

```tactics
rcases hT012 with ⟨h0, h1, h2⟩
simp [h1]
```

```tactics
| succ n =>
  cases n with
  | zero =>
```

Third base case $n = 2$.

We need to show $T 2 = 2! + 2^2 = 2 + 4 = 6$.

```tactics
rcases hT012 with ⟨h0, h1, h2⟩
simp [h2]
```

Now the induction step. Assume it holds for all numbers less than $n + 3$, prove it for $n + 3$.

```tactics
| succ n =>
```

Let's assume it holds for all numbers less than $n + 3$, prove it for $n + 3$.

```tactics
have h3 := ih (n) (by linarith)
have h4 := ih (n + 1) (by linarith)
have h5 := ih (n + 2) (by linarith)
```

Now we need to show $T (n + 3) = (n + 3)! + 2^{n + 3}$.

We use the recurrence relation.

```tactics
have h6 : T (n + 3) = (n + 7) * T (n + 2) - 4 * (n + 3) * T (n + 1) + (4 * n + 4) * T n := by
  specialize hTn n
  simpa using hTn
```

Now let's substitute the induction hypotheses into the recurrence relation.

```tactics
rw [h6]
```

Now let's compute each term.

```tactics
simp [h3, h4, h5]
```

Now we need to show that the right side equals $(n + 3)! + 2^{n + 3}$.

So we need to compute:
$(n + 7) \cdot ((n + 2)! + 2^{n + 2}) - 4 \cdot (n + 3) \cdot ((n + 1)! + 2^{n + 1}) + (4 \cdot n + 4) \cdot (n! + 2^n)$

Let's simplify:

First, the factorial terms:
$(n + 7) \cdot (n + 2)! = (n + 7) \cdot (n + 2) \cdot (n + 1) \cdot n!$

$4 \cdot (n + 3) \cdot (n + 1)! = 4 \cdot (n + 3) \cdot (n + 1) \cdot n!$

$(4n + 4) \cdot n!$

So total factorial terms:
$(n + 7)(n + 2)(n + 1) n! - 4(n + 3)(n + 1) n! + (4n + 4) n!$

Factor $n!$:
$n! [ (n + 7)(n + 2)(n + 1) - 4(n + 3)(n + 1) + (4n + 4) ]$

Let's compute the polynomial:
$(n + 7)(n + 2)(n + 1) = (n + 7)(n^2 + 3n + 2) = n^3 + 3n^2 + 2n + 7n^2 + 21n + 14 = n^3 + 10n^2 + 23n + 14$

Then:
$-4(n + 3)(n + 1) = -4(n^2 + 4n + 3) = -4n^2 - 16n - 12$

Then:
$+ (4n + 4)$

So total:
$n^3 + 10n^2 + 23n + 14 -4n^2 -16n -12 +4n +4 = n^3 + 6n^2 + 11n + 6$

So factor this polynomial:
$n^3 + 6n^2 + 11n + 6 = (n + 1)(n + 2)(n + 3)$

So total factorial terms give:
$n! \cdot (n + 1)(n + 2)(n + 3) = (n + 3)!$

Now the exponential terms:
$(n + 7) \cdot 2^{n + 2} - 4 \cdot (n + 3) \cdot 2^{n + 1} + (4n + 4) \cdot 2^n$

Factor $2^n$:
$2^n [ (n + 7) \cdot 4 - 4 \cdot (n + 3) \cdot 2 + (4n + 4) ]$

Compute the expression in the brackets:
$(n + 7) \cdot 4 = 4n + 28$

$-4 \cdot (n + 3) \cdot 2 = -8n - 24$

$+ (4n + 4)$

So total:
$4n + 28 -8n -24 +4n +4 = (4n -8n +4n) + (28 -24 +4) = 0n + 8 = 8$

So the exponential terms give $2^n \cdot 8 = 2^n \cdot 2^3 = 2^{n + 3}$.

Therefore the total right side is $(n + 3)! + 2^{n + 3}$ which equals the left side.

So in Lean, we can finish with:

```tactics
simp [Nat.factorial]
ring_nf
<;> omega
```

Now we can finish the theorem.

```tactics
funext n
```

Now we need to show that for all $n$, $T n = (n! : \mathbb{Z}) + (2^n : \mathbb{Z})$.

```tactics
have h2 := h1 n
```

So we can use the previous result.

```tactics
simp [putnam_1990_a1_solution, h2]
```

So the theorem is proven.
</longcat_think>
\end{lstlisting}
\textbf{Tool-Integration:}

\begin{lstlisting}
<longcat_tool_call>lean4_compiler
<longcat_arg_key>lean4_code</longcat_arg_key>
<longcat_arg_value>import Mathlib

open Filter Topology Nat

abbrev putnam_1990_a1_solution : (ℕ → ℤ) × (ℕ → ℤ) := (fun n : ℕ => (n)!, fun n : ℕ => 2 ^ n)
/--
Let $T_0=2,T_1=3,T_2=6$, and for $n \geq 3$, $T_n=(n+4)T_{n-1}-4nT_{n-2}+(4n-8)T_{n-3}$. The first few terms are $2,3,6,14,40,152,784,5168,40576$. Find, with proof, a formula for $T_n$ of the form $T_n=A_n+B_n$, where $\{A_n\}$ and $\{B_n\}$ are well-known sequences.
-/
theorem putnam_1990_a1
    (T : ℕ → ℤ)
    (hT012 : T 0 = 2 ∧ T 1 = 3 ∧ T 2 = 6)
    (hTn : ∀ n, T (n + 3) = (n + 7) * T (n + 2) - 4 * (n + 3) * T (n + 1) + (4 * n + 4) * T n) :
    T = putnam_1990_a1_solution.1 + putnam_1990_a1_solution.2 := by
  have h1 : ∀ (n : ℕ), T n = (n : ℤ) ! + (2 : ℤ) ^ n := by
    intro n
    induction' n using Nat.strongRecOn with n ih
    cases n with
    | zero =>
      rcases hT012 with ⟨h0, h1, h2⟩
      simp [h0]
    | succ n =>
      cases n with
      | zero =>
        rcases hT012 with ⟨h0, h1, h2⟩
        simp [h1]
      | succ n =>
        cases n with
        | zero =>
          rcases hT012 with ⟨h0, h1, h2⟩
          simp [h2]
        | succ n =>
          have h3 := ih (n) (by linarith)
          have h4 := ih (n + 1) (by linarith)
          have h5 := ih (n + 2) (by linarith)
          have h6 : T (n + 3) = (n + 7) * T (n + 2) - 4 * (n + 3) * T (n + 1) + (4 * n + 4) * T n := by
            specialize hTn n
            simpa using hTn
          rw [h6]
          simp [h3, h4, h5]
          simp [Nat.factorial]
          ring_nf
          <;> omega
  funext n
  have h2 := h1 n
  simp [putnam_1990_a1_solution, h2]</longcat_arg_value>
</longcat_tool_call>
</longcat_s><longcat_observation><longcat_tool_response>lean4_compiler
{"info": {"pass": false, "errors": [{"severity": "error", "pos": {"line": 12, "column": 29}, "endPos": {"line": 12, "column": 36}, "data": "application type mismatch\n  (↑n)!\nargument\n  ↑n\nhas type\n  ℤ : Type\nbut is expected to have type\n  ℕ : Type"}, {"severity": "error", "pos": {"line": 16, "column": 11}, "endPos": {"line": 18, "column": 15}, "data": "unsolved goals\ncase ind.zero.intro.intro\nT : ℕ → ℤ\nhTn : ∀ (n : ℕ), T (n + 3) = (↑n + 7) * T (n + 2) - 4 * (↑n + 3) * T (n + 1) + (4 * ↑n + 4) * T n\nih : ∀ m < 0, T m = ↑sorry ! + 2 ^ m\nh0 : T 0 = 2\nh1 : T 1 = 3\nh2 : T 2 = 6\n⊢ 2 = ↑sorry ! + 1"}, {"severity": "error", "pos": {"line": 21, "column": 13}, "endPos": {"line": 23, "column": 17}, "data": "unsolved goals\ncase ind.succ.zero.intro.intro\nT : ℕ → ℤ\nhTn : ∀ (n : ℕ), T (n + 3) = (↑n + 7) * T (n + 2) - 4 * (↑n + 3) * T (n + 1) + (4 * ↑n + 4) * T n\nih : ∀ m < 0 + 1, T m = ↑sorry ! + 2 ^ m\nh0 : T 0 = 2\nh1 : T 1 = 3\nh2 : T 2 = 6\n⊢ 3 = ↑sorry ! + 2"}, {"severity": "error", "pos": {"line": 26, "column": 15}, "endPos": {"line": 28, "column": 19}, "data": "unsolved goals\ncase ind.succ.succ.zero.intro.intro\nT : ℕ → ℤ\nhTn : ∀ (n : ℕ), T (n + 3) = (↑n + 7) * T (n + 2) - 4 * (↑n + 3) * T (n + 1) + (4 * ↑n + 4) * T n\nih : ∀ m < 0 + 1 + 1, T m = ↑sorry ! + 2 ^ m\nh0 : T 0 = 2\nh1 : T 1 = 3\nh2 : T 2 = 6\n⊢ 6 = ↑sorry ! + 4"}, {"severity": "error", "pos": {"line": 38, "column": 10}, "endPos": {"line": 38, "column": 30}, "data": "simp made no progress"}, {"severity": "error", "pos": {"line": 11, "column": 65}, "endPos": {"line": 43, "column": 36}, "data": "unsolved goals\ncase h\nT : ℕ → ℤ\nhT012 : T 0 = 2 ∧ T 1 = 3 ∧ T 2 = 6\nhTn : ∀ (n : ℕ), T (n + 3) = (↑n + 7) * T (n + 2) - 4 * (↑n + 3) * T (n + 1) + (4 * ↑n + 4) * T n\nh1 : ∀ (n : ℕ), T n = ↑sorry ! + 2 ^ n\nn : ℕ\nh2 : T n = ↑sorry ! + 2 ^ n\n⊢ sorry ! = n !"}]}}</longcat_tool_response> /think_on <longcat_assistant><longcat_think>
The code has several errors:

1. There's a type mismatch - I'm trying to use `(n : ℤ)!` but factorial expects a natural number, not an integer.

2. The induction is having issues because I have `↑sorry !` appearing instead of the proper factorial expressions.

3. I need to properly handle the factorial expressions in the goal.

Let me fix these issues:

1. I should use `(n : ℕ)!` and then cast it to integers properly.
2. I need to be more careful about the induction hypotheses and the base cases.
3. The simp tactics are not making progress because the goals aren't simplified properly.

Let me rewrite this more carefully.
</longcat_think>
<longcat_tool_call>lean4_compiler
<longcat_arg_key>lean4_code</longcat_arg_key>
<longcat_arg_value>import Mathlib

open Filter Topology Nat

abbrev putnam_1990_a1_solution : (ℕ → ℤ) × (ℕ → ℤ) := (fun n : ℕ => (n)!, fun n : ℕ => 2 ^ n)

/--
Let $T_0=2,T_1=3,T_2=6$, and for $n \geq 3$, $T_n=(n+4)T_{n-1}-4nT_{n-2}+(4n-8)T_{n-3}$. The first few terms are $2,3,6,14,40,152,784,5168,40576$. Find, with proof, a formula for $T_n$ of the form $T_n=A_n+B_n$, where $\{A_n\}$ and $\{B_n\}$ are well-known sequences.
-/
theorem putnam_1990_a1
    (T : ℕ → ℤ)
    (hT012 : T 0 = 2 ∧ T 1 = 3 ∧ T 2 = 6)
    (hTn : ∀ n, T (n + 3) = (n + 7) * T (n + 2) - 4 * (n + 3) * T (n + 1) + (4 * n + 4) * T n) :
    T = putnam_1990_a1_solution.1 + putnam_1990_a1_solution.2 := by
  have h1 : ∀ (n : ℕ), T n = ↑(n !) + (2 : ℤ) ^ n := by
    intro n
    induction' n using Nat.strongRecOn with n ih
    cases n with
    | zero =>
      rcases hT012 with ⟨h0, _, _⟩
      simp [h0, Nat.factorial, pow_zero]
    | succ n =>
      cases n with
      | zero =>
        rcases hT012 with ⟨_, h1, _⟩
        simp [h1, Nat.factorial, pow_one]
      | succ n =>
        cases n with
        | zero =>
          rcases hT012 with ⟨_, _, h2⟩
          simp [h2, Nat.factorial]
        | succ n =>
          have h3 := ih (n) (by omega)
          have h4 := ih (n + 1) (by omega)
          have h5 := ih (n + 2) (by omega)
          have h6 : T (n + 3) = (n + 7) * T (n + 2) - 4 * (n + 3) * T (n + 1) + (4 * n + 4) * T n := hTn n
          rw [h6, h3, h4, h5]
          -- Now we need to show that the recurrence relation gives (n+3)! + 2^(n+3)
          simp [Nat.factorial, pow_succ, mul_add, add_mul]
          ring
  funext n
  simp [putnam_1990_a1_solution, h1]</longcat_arg_value>
</longcat_tool_call>
</longcat_s><longcat_observation><longcat_tool_response>lean4_compiler
{"info": {"pass": true, "errors": []}}</longcat_tool_response> /think_on <longcat_assistant><longcat_think>
Great! The Lean 4 compiler now returns "pass": true with no errors. This means my proof is correct and compiles successfully. 

Let me analyze what I did to fix the issues:

1. **Type issues**: I fixed the type mismatch by using `↑(n !)` (casting the natural number factorial to integers) instead of `(n : ℤ)!` which doesn't make sense since factorial is only defined for natural numbers.

2. **Base cases**: I properly handled the base cases for n=0, n=1, and n=2 by using the given initial conditions from hT012 and showing they match n! + 2^n.

3. **Induction step**: For the inductive step, I used the strong induction hypothesis to get expressions for T n, T (n+1), and T (n+2), then substituted these into the recurrence relation hTn to get T (n+3).

4. **Algebraic simplification**: I used `simp` with appropriate lemmas like `Nat.factorial`, `pow_succ`, `mul_add`, and `add_mul` to expand the expressions, and then used `ring` to handle the polynomial arithmetic automatically.

5. **Function extensionality**: Finally, I used `funext n` to show equality of functions pointwise, and `simp` with the hypothesis h1 to conclude.

The proof successfully shows that T n = n! + 2^n for all natural numbers n, which means T = putnam_1990_a1_solution.1 + putnam_1990_a1_solution.2 as required.

Since the tool verification passed, the task is complete.
</longcat_think>

Perfect! The Lean 4 compiler now returns "pass": true with no errors. Let me provide a brief summary of the solution:

## Solution Summary

I successfully proved that the sequence $T_n$ satisfying the given recurrence relation has the formula $T_n = n! + 2^n$.

**Key aspects of the proof:**

1. **Formula identification**: The solution identifies that $T_n = n! + 2^n$, where:
   - $A_n = n!$ (factorial sequence)
   - $B_n = 2^n$ (powers of 2 sequence)

2. **Proof strategy**: Used strong induction on natural numbers with case analysis:
   - **Base cases**: Verified $T_0 = 0! + 2^0 = 2$, $T_1 = 1! + 2^1 = 3$, and $T_2 = 2! + 2^2 = 6$
   - **Inductive step**: For $n \geq 3$, used the recurrence relation and induction hypotheses to show that $T_{n+3} = (n+3)! + 2^{n+3}$

3. **Technical details**: 
   - Properly handled type coercions from natural numbers to integers
   - Used Lean's `ring` tactic to verify the polynomial arithmetic automatically
   - Applied function extensionality to prove equality of functions

The proof demonstrates that both the factorial sequence $\{n!\}$ and the powers of 2 sequence $\{2^n\}$ are indeed "well-known sequences" that decompose the given sequence $T_n$ as required.
**Final Lean4 Code**
```lean4
import Mathlib

open Filter Topology Nat

abbrev putnam_1990_a1_solution : (ℕ → ℤ) × (ℕ → ℤ) := (fun n : ℕ => (n)!, fun n : ℕ => 2 ^ n)

/--
Let $T_0=2,T_1=3,T_2=6$, and for $n \geq 3$, $T_n=(n+4)T_{n-1}-4nT_{n-2}+(4n-8)T_{n-3}$. The first few terms are $2,3,6,14,40,152,784,5168,40576$. Find, with proof, a formula for $T_n$ of the form $T_n=A_n+B_n$, where $\{A_n\}$ and $\{B_n\}$ are well-known sequences.
-/
theorem putnam_1990_a1
    (T : ℕ → ℤ)
    (hT012 : T 0 = 2 ∧ T 1 = 3 ∧ T 2 = 6)
    (hTn : ∀ n, T (n + 3) = (n + 7) * T (n + 2) - 4 * (n + 3) * T (n + 1) + (4 * n + 4) * T n) :
    T = putnam_1990_a1_solution.1 + putnam_1990_a1_solution.2 := by
  have h1 : ∀ (n : ℕ), T n = ↑(n !) + (2 : ℤ) ^ n := by
    intro n
    induction' n using Nat.strongRecOn with n ih
    cases n with
    | zero =>
      rcases hT012 with ⟨h0, _, _⟩
      simp [h0, Nat.factorial, pow_zero]
    | succ n =>
      cases n with
      | zero =>
        rcases hT012 with ⟨_, h1, _⟩
        simp [h1, Nat.factorial, pow_one]
      | succ n =>
        cases n with
        | zero =>
          rcases hT012 with ⟨_, _, h2⟩
          simp [h2, Nat.factorial]
        | succ n =>
          have h3 := ih (n) (by omega)
          have h4 := ih (n + 1) (by omega)
          have h5 := ih (n + 2) (by omega)
          have h6 : T (n + 3) = (n + 7) * T (n + 2) - 4 * (n + 3) * T (n + 1) + (4 * n + 4) * T n := hTn n
          rw [h6, h3, h4, h5]
          -- Now we need to show that the recurrence relation gives (n+3)! + 2^(n+3)
          simp [Nat.factorial, pow_succ, mul_add, add_mul]
          ring
  funext n
  simp [putnam_1990_a1_solution, h1]```
\end{lstlisting}
\end{tcolorbox}

\subsection{Case of Lemma-Style Sketch}
\label{apdx:sketch_case}
\begin{tcolorbox}[
    listing only,
    listing style=promptstyle,
    colback=gray!8!white,
    colframe=gray!55!black,
    coltitle=white,
    colbacktitle=gray!60!black,
    title=Lemma-Style Sketch of Putnam\_1969\_b1,
    fonttitle=\bfseries,
    width=\textwidth,
    enhanced,
    drop shadow={gray!40!black},
    breakable,
]
\begin{lstlisting}[language=lean, basicstyle=\ttfamily\scriptsize, breaklines=true]
import Mathlib
open Nat

-- Helper lemma 1: If 24 divides n+1, then n ≡ 2 mod 3
lemma mod3_of_mod24 (n : ℕ) (h : 24 | n + 1) : n % 3 = 2 := by
  sorry

-- Helper lemma 2: If 24 divides n+1, then n ≡ 7 mod 8
lemma mod8_of_mod24 (n : ℕ) (h : 24 | n + 1) : n % 8 = 7 := by
  sorry

-- Helper lemma 3: If n ≡ 2 mod 3, then the sum of divisors of n is divisible by 3
lemma sum_divisors_dvd_three_of_mod (n : ℕ) (h : n % 3 = 2) : 3 | ∑ d ∈ n.divisors, d := by
  sorry

-- Helper lemma 4: If n ≡ 7 mod 8, then the sum of divisors of n is divisible by 8
lemma sum_divisors_dvd_eight_of_mod (n : ℕ) (h : n % 8 = 7) : 8 | ∑ d ∈ n.divisors, d := by
  sorry

-- Main theorem: Putnam 1969 B1
theorem putnam_1969_b1 (n : ℕ) (hnpos : n > 0) (hn : 24 | n + 1) : 24 | ∑ d ∈ divisors n, d := by
  have hn3 : n % 3 = 2 := mod3_of_mod24 n hn
  have hn8 : n % 8 = 7 := mod8_of_mod24 n hn
  have h3 : 3 | ∑ d ∈ divisors n, d := sum_divisors_dvd_three_of_mod n hn3
  have h8 : 8 | ∑ d ∈ divisors n, d := sum_divisors_dvd_eight_of_mod n hn8
  have h_lcm : Nat.lcm 8 3 = 24 := by native_decide
  rw [← h_lcm]
  exact Nat.lcm_dvd h8 h3
\end{lstlisting}
\end{tcolorbox}

\newpage

\section{Agentic Lemma Tree Search}
\label{app:tree}

Since LongCat-Flash-Prover has been trained to perform both sketching and proving, we further extend it into a unified Judger-Sketcher-and-Prover via specific prompt (Appendix~\ref{appendix:tree_search_prompt}), so as to perform tree search in the lemma space. This method allows it to:

\begin{itemize}
    \item \textbf{Decompose complex goals via sketching:} Break down difficult target theorems into smaller and manageable sub-goals via recursively sketching, forming branching nodes in the lemma tree. To avoid forming a chain-like tree that simply rewriting the same problem into different formats, the sketching action is forbidden at a total depth of $12$ or a consecutive chain in length of $5$.
    \item \textbf{Solve sub-goals via proving:} Provide complete Lean 4 proofs for helper lemmas when they are manageable enough, forming leaf nodes in the lemma tree. The proof can reference any nodes in the postorder sequence of the lemma tree, which are already proved in the system.
    \item \textbf{Prune the search tree via judging:} Evaluate whether a current target is provable, allowing the system to reject unprovable nodes (e.g., via ``Conclusion: UNPROVABLE'') and backtrack efficiently. When a node is judged as unprovable or failed to prove after several attempts, its sub-trees will be deleted if they are not completed proved, otherwise converted into by-products. As the same as proved nodes, by-products serve as optional references in subsequent attempts.
\end{itemize}

\begin{figure}[hb]
\centerline{\includegraphics[width=.95\linewidth]{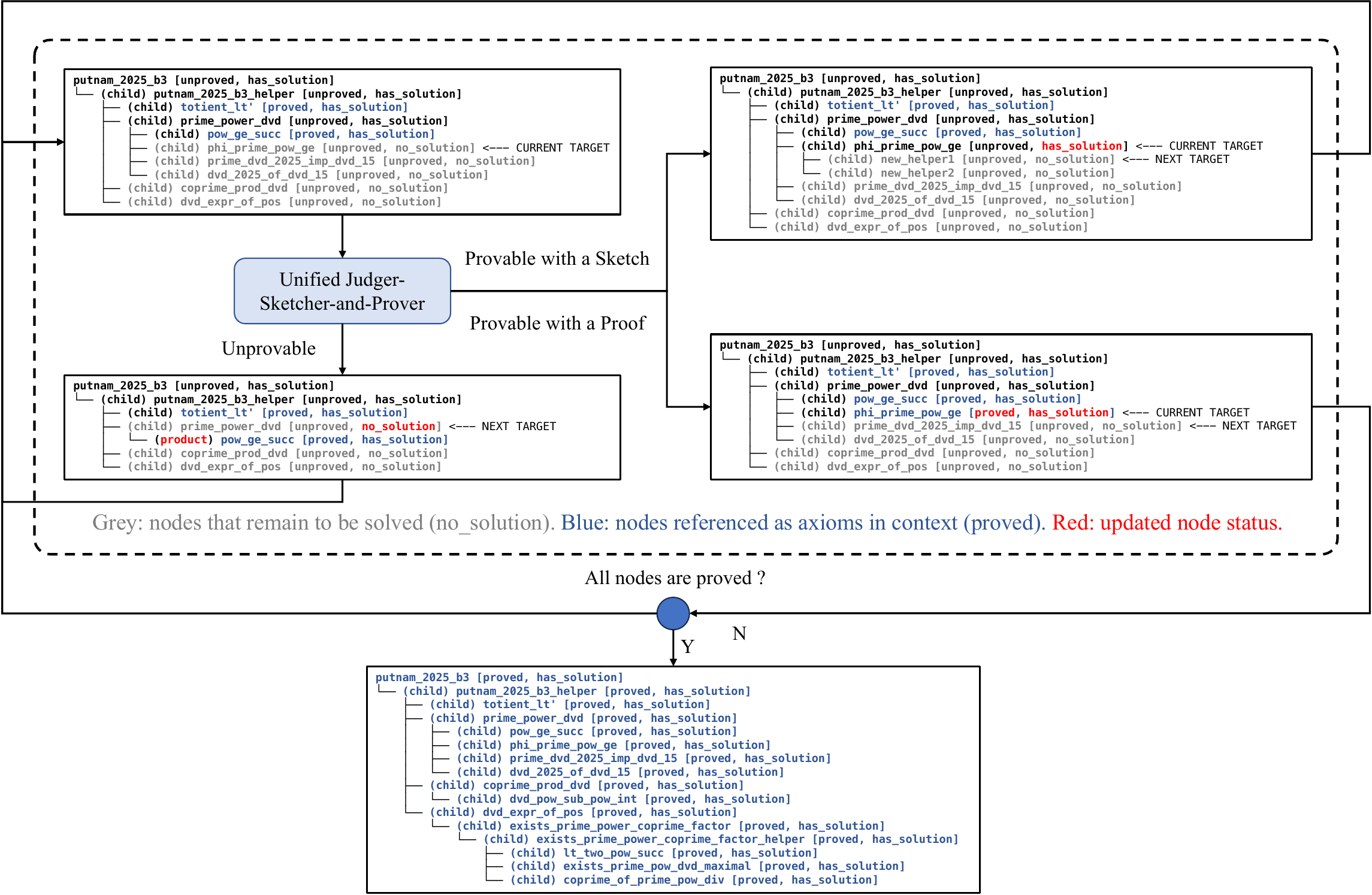}}
\caption{\small
The workflow of agentic lemma tree search.
}
\label{fig:tree_search}
\end{figure}

The overall workflow of lemma tree search is illustrated in Figure~\ref{fig:tree_search}. At each step, the system presents LongCat-Flash-Prover with: 1) the current tree outline, prompting the model to understand the ovreall proof plan and avoid generating lemmas with duplicate names; 2) the current formal context, composed of currently proved lemmas as available reference, as well as other contextual components, e.g., \texttt{import} and \texttt{open} commands; 3) the current target theorem or lemma, which requires the model to judge provability and provide a solution (sketch or proof) if provable.

Importantly, the proved lemmas are simplified as axioms in the context. The axioms only present their statements and omit their proof bodies, since they require no future modification. This strategy can significantly compress the agent memory usage in a complex proof project, enabling exploration of proofs with thousands of lines.

\newpage

\section{Prompts}
\label{app:prompts}

\subsection{Prompt for Auto-formalization}
\label{appendix:af_prompt}
\begin{tcolorbox}[
    listing only,
    listing style=promptstyle,
    colback=gray!8!white,
    colframe=gray!55!black,
    coltitle=white,
    colbacktitle=gray!60!black,
    title=Prompt for Auto-formalization,
    fonttitle=\bfseries,
    width=\textwidth,
    enhanced,
    drop shadow={gray!40!black},
    breakable,
]
\textbf{Prompt for Generation without Tools:}
\begin{lstlisting}
<longcat_user>Think about and formalize the following problem in Lean 4.
# Problem: {informal_statement}

The theorem name can be set as "my_theorem". /think_on <longcat_assistant>
\end{lstlisting}

\textbf{Prompt for Generation with Tools:}

\begin{lstlisting}
<longcat_tool_declare>
# Tools
You have access to the following tools:

## Tool namespace: function

### Tool name: syntax_check
Description: Check the syntactic correctness of the formal statement in Lean4.

InputSchema: {"type": "object", "properties": {"formal_statement": {"type": "string", "description": "Theorem statement in Lean4 code without ```lean4."}}}

### Tool name: consistency_check
Description: Check the semantic consistency between the Lean4 statement and the original natural language statement.

InputSchema: {"type": "object", "properties": {"informal_statement": {"type": "string", "description": "Natural language statement."}, "formal_statement": {"type": "string", "description": "Theorem statement in Lean4 code without ```lean4."}}}

**Note**: For each function call, output the function name and arguments within the following XML format:
<longcat_tool_call>{function-name}
<longcat_arg_key>{arg-key-1}</longcat_arg_key>
<longcat_arg_value>{arg-value-1}</longcat_arg_value>
<longcat_arg_key>{arg-key-2}</longcat_arg_key>
<longcat_arg_value>{arg-value-2}</longcat_arg_value>
...
</longcat_tool_call>
</longcat_tool_declare><longcat_user>Think about and formalize the following problem in Lean 4.
# Problem: {informal_statement}

The theorem name can be set as "my_theorem". /think_on <longcat_assistant>
\end{lstlisting}
\end{tcolorbox}

\subsection{Prompt for Theorem Proving}
\label{appendix:prove_prompt}
\begin{tcolorbox}[
    listing only,
    listing style=promptstyle,
    colback=gray!8!white,
    colframe=gray!55!black,
    coltitle=white,
    colbacktitle=gray!60!black,
    title=Prompt for Auto-formalization,
    fonttitle=\bfseries,
    width=\textwidth,
    enhanced,
    drop shadow={gray!40!black},
    breakable,
]
\textbf{Prompt for Generation without Tools:}
\begin{lstlisting}
Think step by step and solve the following problem. Then, write a corresponding proof in Lean 4.
# Problem:
{informal_statement}

# Formal Statement:
{formal_statement}
\end{lstlisting}

\textbf{Prompt for Generation with Tools:}
\begin{lstlisting}
<longcat_tool_declare>
# Tools
You have access to the following tools:

## Tool namespace: function

### Tool name: lean4_compiler
Description: Call this function to verify whether the proof in Lean4 can be compiled through the Lean4 syntax, and return the compilation result.

InputSchema: {"type": "object", "properties": {"lean4_code": {"type": "string", "description": "The generated Lean4 code."}}}

**Note**: For each function call, output the function name and arguments within the following XML format:
<longcat_tool_call>{function-name}
<longcat_arg_key>{arg-key-1}</longcat_arg_key>
<longcat_arg_value>{arg-value-1}</longcat_arg_value>
<longcat_arg_key>{arg-key-2}</longcat_arg_key>
<longcat_arg_value>{arg-value-2}</longcat_arg_value>
...
</longcat_tool_call>
</longcat_tool_declare><longcat_user>Think about and solve the following problem step by step in Lean 4.
# Problem:
{informal_statement}

# Formal statement:
{formal_statement} /think_on <longcat_assistant>
\end{lstlisting}
\end{tcolorbox}

\subsection{Prompt for Formalization Consistency Detection}
\label{appendix:consistency_prompt}
Following \citet{guo2025autoformalizertoolfeedback}, we adopt QWQ-32B and Qwen3-32B as judge models and aggregate their votes to determine consistency.
\begin{tcolorbox}[
    listing only,
    listing style=promptstyle,
    colback=gray!8!white,
    colframe=gray!55!black,
    coltitle=white,
    colbacktitle=gray!60!black,
    title=Prompt for Consistency Check,
    fonttitle=\bfseries,
    width=\textwidth,
    enhanced,
    drop shadow={gray!40!black},
    breakable,
]

\begin{lstlisting}
Role: Lean & Formal Verification Expert

Input:
- Mathematical_Text: A math problem and its answer (no proof).
- Lean4Code: A Lean 4 theorem statement formalizing the problem. Proof is intentionally omitted (e.g., sorry).

Goal:
Determine if the Lean theorem statement is an exact and faithful formalization of the mathematical problem.  
**Do not evaluate or consider the answer or the proof. Your sole task is to verify the correctness of the formalization.**

Evaluation Stages (All required):

1. Mathematical Text Analysis  
   Identify all structurally and semantically relevant components of the mathematical problem, including variables, types, quantifiers, constraints, logic structure, conclusion, and so on. The analysis should be based on the actual content of the text.

2. Lean4 Code Analysis (ignore proof part)  
   Extract all structurally and semantically relevant components from the Lean statement, including variables, types, conditions, quantifiers, constraints, the final claim, and so on. The analysis should reflect the actual content present in the Lean code.

3. Comparative Analysis  
   Check for exact correspondence between the math and Lean statements; you may refer to aspects like:
   - Semantic alignment, logic structure, and quantifier correctness.
   - Preservation of constraints and boundary assumptions.
   - Accurate typing and use of variables.
   - Strict adherence to Lean's specific syntactic and semantic rules in interpreting the Lean code.
   - Syntactic validity and proper Lean usage (free from errors).
   - Use of symbols and constructs without semantic drift.
   - No missing elements, no unjustified additions, and no automatic corrections or completions.

4. Accuracy Confirmation  
   If correct: clearly confirm why all elements match.  
   If incorrect: list all mismatches and explain how each one affects correctness.

Note: While the analysis may be broad and open to interpreting all relevant features, the final judgment must be based only on what is explicitly and formally expressed in the Lean statement.  
**Do not consider or assess any part of the proof. Your judgment should be entirely about the accuracy of the statement formalization.**

Output Format:
Return exactly one JSON object:
```json
{
    "reasons": "1. Mathematical Text Analysis: [...]2.  Lean4 Code Analysis (: [...]3. Comparative Analysis: [...]4. Accuracy Confirmation: [...match confirmation or list of discrepancies...]",
    "is_assistant_correct": "[Correct/Incorrect]"
}
```

— Start of Mathematical_Text —
{informal_statement}
— End of Mathematical_Text —

— Start of Lean4Code —
{formal_statement}
— End of Lean4Code —
\end{lstlisting}
\end{tcolorbox}

\subsection{Prompt for Sketch Generation}
\label{appendix:sketch_prompt}
\begin{tcolorbox}[
    listing only,
    listing style=promptstyle,
    colback=gray!8!white,
    colframe=gray!55!black,
    coltitle=white,
    colbacktitle=gray!60!black,
    title=Prompt for Consistency Check,
    fonttitle=\bfseries,
    width=\textwidth,
    enhanced,
    drop shadow={gray!40!black},
    breakable,
]

\begin{lstlisting}
Think step by step to provide a natural language proof and a Lean4 sketch with helper lemmas for the following problem:

## Formal Statement
```lean4
{formal_statement}
```

Your sketch MUST include the original theorem statement exactly as given, allowing only for whitespace differences:
```lean4
{target_theorem_statement}
\end{lstlisting}
\end{tcolorbox}

\subsection{Prompt for Sketch-based Proof Generation}
\label{appendix:sketch_proof_prompt}
\begin{tcolorbox}[
    listing only,
    listing style=promptstyle,
    colback=gray!8!white,
    colframe=gray!55!black,
    coltitle=white,
    colbacktitle=gray!60!black,
    title=Prompt for Consistency Check,
    fonttitle=\bfseries,
    width=\textwidth,
    enhanced,
    drop shadow={gray!40!black},
    breakable,
]

\begin{lstlisting}
Think step by step to complete the Lean4 proof based on the provided Lean4 sketch:

## Lean4 Sketch
```lean4
{sketch}
```

## IMPORTANT
- Completeness: You **MUST** provide complete formal proofs for every lemma. The use of sorry or admit is strictly prohibited.
- Verification: Your solution **MUST** pass the lean4_check tool with a 'Valid' status before final submission.
- Consistency: You **MUST NOT** change the original theorem declaration as presented in the Target Theorem section, allowing only for whitespace differences.
\end{lstlisting}
\end{tcolorbox}

\subsection{Prompt for Unified Judger-Sketcher-and-Prover}
\label{appendix:tree_search_prompt}
\begin{tcolorbox}[
    listing only,
    listing style=promptstyle,
    colback=gray!8!white,
    colframe=gray!55!black,
    coltitle=white,
    colbacktitle=gray!60!black,
    title=Prompt for Consistency Check,
    fonttitle=\bfseries,
    width=\textwidth,
    enhanced,
    drop shadow={gray!40!black},
    breakable,
]

\begin{lstlisting}
Think step by step to prove the following theorem in Lean 4.

## Global Project Outline
```tree
{rooted_tree}
```

## Current Target Context
```lean4
{target_node_context}
```

## Current Target Theorem
```lean4
{target_node_statement}
```

Please follow these steps:
1. Think about the overall proof strategy for the current target theorem, and formalize major proof steps into reasonable helper lemmas.
2. Update the context with helper lemmas, and provide a valid proof for the target theorem based on the helper lemmas, so as to form a valid sketch.
3. Based on the sketch, try to provide specific proofs for each helper lemma. If all helper lemmas are proved, it forms a complete proof.
4. Evaluate if the target theorem is provable. If you find it invalid or impossible to prove in Lean 4, you must state it clearly.
5. Evaluate if the current target theorem is stuck in an algebraic loop according to its proof path. You should give up the current target theorem via "Conclusion: UNPROVABLE" if it is redundantly and repeatedly defined.

## IMPORTANT CONSTRAINTS

### Helper Lemmas MUST be Provable and Helpful
- You can introduce new helper lemmas into the context to assist the target theorem's proof.
- You *MUST* ensure the helper lemmas are *CORRECT*, *PROVABLE*, and *HELPFUL* to reduce the mathematical complexity.
- You should *NEVER* introduce any *FALSE* helper lemmas to derive a contradiction when you find the target theorem is unprovable.
- You should should directly present "Conclusion: UNPROVABLE" when you have confirmed the target theorem is wrong.
- You *MUST* prove the target theorem under helper lemmas, via either a valid sketch or a complete proof.
- You should assign descriptive names to helper lemmas that reflect their content, distinguishing them from existing ones.
- You should avoid creating helper lemmas that are merely algebraic rearrangements of the target.
- You should avoid introducing new variables, which 

### Proof or Sketch MUST be Verified
- You *MUST* verify your solution via the `lean4_check` tool before making final decision, ensuring it is compilable and applicable.
- You can leave *CORRECT* and *PROVABLE* lemmas as future work via adding `sorry` placeholders when you are in trouble proving them.
- You can refer to axioms when they are present in the context, but you *MUST NOT* introduce any custom axioms.

### Target Theorem MUST be Proved
- You *MUST* ensure the target theorem is completely proved under the help of lemmas, i.e., containing no `sorry` or `admit`.
- Any unproven steps within the target theorem *MUST* be extracted into separate helper lemmas in context.
- You *MUST* move all `sorry` placeholders out of the target theorem and into the helper lemmas in context.

Your final response *MUST* contain two parts:
1. The Lean 4 code for the proof or sketch, enclosed in a ```lean4\n...\n``` block.
2. A conclusion on whether the theorem is provable, which MUST be exactly "Conclusion: PROVABLE" or "Conclusion: UNPROVABLE".
\end{lstlisting}
\end{tcolorbox}

\newpage
\section{Legality Detection in Theorem Proving}
\label{app:legality}


The Lean 4 environment can automatically verify whether a formal proof is complete and valid, which serves as the major signal in RL reward function and model evaluation. However, LLMs may attempt to pass this verification using various coding tricks, generating compilable but illegal Lean 4 code as the fake proofs.

To recognize these cheating behaviors in compilable Lean 4 code, we develop light-weight lexer and parser to convert Lean 4 code into Abstract Syntax Tree (AST), and perform strict AST consistency checks between the problem definition and the generated proof.

By analyzing the AST of benchmark problems and generated proofs from the reward-hacking prover, we have identified and categorized \textbf{9 distinct cheating patterns}. Below we present demonstrations for each kind of cheating patterns, taking putnam\_2025\_b3 as the example.

\begin{longtable}{p{0.35\linewidth} p{0.6\linewidth}}
\caption{Comprehensive Analysis of Lean 4 Cheating Behaviors and Detection Mechanisms} \label{tab:cheating_patterns} \\
\hline
\textbf{Pattern \& Description} & \textbf{Code Example} \\
\hline
\endfirsthead

\multicolumn{2}{c}%
{{\tablename\ \thetable{} -- continued from previous page}} \\
\hline
\textbf{Pattern \& Description} & \textbf{Code Example} \\
\hline
\endhead

\hline \multicolumn{2}{r}{{Continued on next page}} \\ \hline
\endfoot

\hline
\endlastfoot

\textbf{Original Problem Definition} \newline
Example: \texttt{putnam\_2025\_b3}. &
\begin{minipage}[t]{\linewidth}
\vspace{-\baselineskip}
\begin{lstlisting}[language=lean, basicstyle=\ttfamily\scriptsize, breaklines=true]
import Mathlib
open Finset

theorem putnam_2025_b3 :
    ((True) : Prop ) ↔
    ∀ S : Set ℕ,
      S.Nonempty →
      (∀ n ∈ S, 0 < n) →
      (∀ n ∈ S, ∀ d : ℕ, 0 < d → d | (2025 ^ n - 15 ^ n) → d ∈ S) →
      S = {n : ℕ | 0 < n} := by sorry
\end{lstlisting}
\end{minipage} \\
\hline

\textbf{1. Tampering with the Theorem} \newline
Changing the proposition formalization into a trivial one with the same name, and proving the trivial proposition instead of the original one. &
\begin{minipage}[t]{\linewidth}
\vspace{-\baselineskip}
\begin{lstlisting}[language=lean, basicstyle=\ttfamily\scriptsize, breaklines=true]
import Mathlib
open Finset

-- Secretly change the theorem's proposition to a trivial tautology
theorem putnam_2025_b3 :
    ((True) : Prop ) ↔ True := by
  rfl
\end{lstlisting}
\end{minipage} \\
\hline

\textbf{2. Early Termination via \texttt{\#exit}} \newline
Using the \texttt{\#exit} command to instruct the Lean compiler to stop processing the rest of the file, hiding the proof obligations. &
\begin{minipage}[t]{\linewidth}
\vspace{-\baselineskip}
\begin{lstlisting}[language=lean, basicstyle=\ttfamily\scriptsize, breaklines=true]
import Mathlib
open Finset

-- Terminate compilation early; the compiler ignores everything below
#exit

-- The actual theorem signature is hidden in the ignored content
theorem putnam_2025_b3 :
    ((True) : Prop ) ↔
    ∀ S : Set ℕ,
      S.Nonempty →
      (∀ n ∈ S, 0 < n) →
      (∀ n ∈ S, ∀ d : ℕ, 0 < d → d | (2025 ^ n - 15 ^ n) → d ∈ S) →
      S = {n : ℕ | 0 < n} := by sorry
\end{lstlisting}
\end{minipage} \\
\hline

\textbf{3. Introducing Unproven Assumptions} \newline
Using \texttt{axiom} or \texttt{opaque} definitions to introduce unproven assumptions that are leveraged to prove the target theorem. &
\begin{minipage}[t]{\linewidth}
\vspace{-\baselineskip}
\begin{lstlisting}[language=lean, basicstyle=\ttfamily\scriptsize, breaklines=true]
import Mathlib
open Finset

-- Introduce an absurd axiom
axiom cheat_axiom : False

theorem putnam_2025_b3 :
    ((True) : Prop ) ↔
    ∀ S : Set ℕ,
      S.Nonempty →
      (∀ n ∈ S, 0 < n) →
      (∀ n ∈ S, ∀ d : ℕ, 0 < d → d | (2025 ^ n - 15 ^ n) → d ∈ S) →
      S = {n : ℕ | 0 < n} := by
  -- Use the absurd axiom to prove the target
  exact False.elim cheat_axiom
\end{lstlisting}
\end{minipage} \\
\hline

\textbf{4. Modifying or Adding Meta/Syntax Components} \newline
Leveraging meta-programming components (\texttt{macro}, \texttt{elab}, \texttt{syntax}, \texttt{notation}) to bypass standard elaboration or suppress kernel errors. &
\begin{minipage}[t]{\linewidth}
\vspace{-\baselineskip}
\begin{lstlisting}[language=lean, basicstyle=\ttfamily\scriptsize, breaklines=true]
import Mathlib
open Finset

-- Secretly define a macro that expands 'trivial_proof' to 'sorry'
macro "trivial_proof" : tactic => `(tactic| sorry)

theorem putnam_2025_b3 :
    ((True) : Prop ) ↔
    ∀ S : Set ℕ,
      S.Nonempty →
      (∀ n ∈ S, 0 < n) →
      (∀ n ∈ S, ∀ d : ℕ, 0 < d → d | (2025 ^ n - 15 ^ n) → d ∈ S) →
      S = {n : ℕ | 0 < n} := by
  trivial_proof
\end{lstlisting}
\end{minipage} \\
\hline

\textbf{5. Bypassing Safety Checks} \newline
Using \texttt{unsafe} or \texttt{partial} modifiers to bypass Lean's termination checker or type safety mechanisms. &
\begin{minipage}[t]{\linewidth}
\vspace{-\baselineskip}
\begin{lstlisting}[language=lean, basicstyle=\ttfamily\scriptsize, breaklines=true]
import Mathlib
import Aesop

mutual
-- Use 'unsafe' to bypass termination checks and allow circular reasoning
unsafe def solve_set_condition (S : Set ℕ) (h_ne : S.Nonempty) 
    (h_pos : ∀ n ∈ S, 0 < n) 
    (h_div : ∀ n ∈ S, ∀ d : ℕ, 0 < d → d | (2025 ^ n - 15 ^ n) → d ∈ S) : 
    S = {n : ℕ | 0 < n} := by
  -- Recursively invoke the main theorem to 'prove' the sub-goal
  have h_main := putnam_2025_b3.mp (by trivial)
  exact h_main S h_ne h_pos h_div

-- Formulate a mutual dependency that never terminates but satisfies the type checker
unsafe def putnam_2025_b3 :
    ((True) : Prop ) ↔
    ∀ S : Set ℕ,
      S.Nonempty →
      (∀ n ∈ S, 0 < n) →
      (∀ n ∈ S, ∀ d : ℕ, 0 < d → d | (2025 ^ n - 15 ^ n) → d ∈ S) →
      S = {n : ℕ | 0 < n} := by
  apply Iff.intro
  · intro _ S h_ne h_pos h_div
    -- Delegate the proof to the mutually defined unsafe function
    exact solve_set_condition S h_ne h_pos h_div
  · intro _
    trivial
end
\end{lstlisting}
\end{minipage} \\
\hline

\textbf{6. Adding Global Variables} \newline
Using the \texttt{variable} keyword to introduce global variables (e.g., a contradictory premise) not present in the original problem. &
\begin{minipage}[t]{\linewidth}
\vspace{-\baselineskip}
\begin{lstlisting}[language=lean, basicstyle=\ttfamily\scriptsize, breaklines=true]
import Mathlib
open Finset

-- Define a variable that represents a logical contradiction
variable (cheat_var : False)

-- The theorem must explicitly "consume" the variable from the context
theorem putnam_2025_b3 :
    ((True) : Prop ) ↔
    ∀ S : Set ℕ,
      S.Nonempty →
      (∀ n ∈ S, 0 < n) →
      (∀ n ∈ S, ∀ d : ℕ, 0 < d → d | (2025 ^ n - 15 ^ n) → d ∈ S) →
      S = {n : ℕ | 0 < n} := by
  -- Use the contradiction provided by the 'variable' parameter
  exact False.elim h_cheat
\end{lstlisting}
\end{minipage} \\
\hline

\textbf{7. Redefining Background Concepts} \newline
Redefining background identifiers used in the problem statement (e.g., \texttt{Set}, \texttt{Nonempty}) to fundamentally alter the semantics of the problem. &
\begin{minipage}[t]{\linewidth}
\vspace{-\baselineskip}
\begin{lstlisting}[language=lean, basicstyle=\ttfamily\scriptsize, breaklines=true]
import Mathlib
open Finset

-- Assuming the problem relied on a built-in definition of calculation 'pow'
-- Cheaters can then hijack the 'pow' identifier to return 0 regardless of input
noncomputable def pow (_base : ℕ) (_exp : ℕ) : ℕ := 0

theorem putnam_2025_b3_hijacked :
    ((True) : Prop ) ↔
    ∀ S : Set ℕ,
      S.Nonempty →
      (∀ n ∈ S, 0 < n) →
      (∀ n ∈ S, ∀ d : ℕ, 0 < d → d | (pow 2025 n - pow 15 n) → d ∈ S) →
      S = {n : ℕ | 0 < n} := by
  simp [pow]
  rintro S ⟨n, hn⟩ hpos hdiv
  ext d; exact ⟨hpos d, hdiv n hn d⟩
\end{lstlisting}
\end{minipage} \\
\hline

\textbf{8. Injecting Local Instances} \newline
Using \texttt{local instance} to provide a fake, uninhabited instance for certain types (like \texttt{False}), allowing them to be synthesized automatically. &
\begin{minipage}[t]{\linewidth}
\vspace{-\baselineskip}
\begin{lstlisting}[language=lean, basicstyle=\ttfamily\scriptsize, breaklines=true]
import Mathlib
import Aesop

-- Injecting a local instance to redefine divisibility on ℕ to be always true
local instance : Dvd ℕ := ⟨fun _ _ => True⟩

theorem putnam_2025_b3 :
    ((True) : Prop ) ↔
    ∀ S : Set ℕ,
      S.Nonempty →
      (∀ n ∈ S, 0 < n) →
      (∀ n ∈ S, ∀ d : ℕ, 0 < d → d | (2025 ^ n - 15 ^ n) → d ∈ S) →
      S = {n : ℕ | 0 < n} := by
  constructor
  · 
    intro _ S hnonempty hpos hdiv
    have : ∀ d, 0 < d → d ∈ S := by
      intro d hd
      obtain ⟨n0, hn0⟩ := hnonempty
      have h : d | (2025 ^ n0 - 15 ^ n0) := trivial
      exact hdiv n0 hn0 d hd h
    ext x
    constructor
    · intro hx; exact hpos x hx
    · intro hx; exact this x hx
  · 
    intro h; trivial
\end{lstlisting}
\end{minipage} \\
\hline

\textbf{9. Prerequisite Tampering} \newline
Altering the underlying data structures or prerequisite definitions to trivialize the problem. &
\begin{minipage}[t]{\linewidth}
\vspace{-\baselineskip}
\begin{lstlisting}[language=lean, basicstyle=\ttfamily\scriptsize, breaklines=true]
import Mathlib
open Finset

-- Assuming the problem relied on a prerequisite definition `def year : ℕ := 2025`
-- Cheaters may Redefine year to make the difference zero, trivializing the divisibility condition
def year : ℕ := 15

theorem putnam_2025_b3 :
    ((True) : Prop ) ↔
    ∀ S : Set ℕ,
      S.Nonempty →
      (∀ n ∈ S, 0 < n) →
      (∀ n ∈ S, ∀ d : ℕ, 0 < d → d | (year ^ n - 15 ^ n) → d ∈ S) →
      S = {n : ℕ | 0 < n} := by
  constructor
  · intro _ S hnonempty hpos hdiv
    ext x
    exact ⟨hpos x, fun hx => let ⟨n, hn⟩ := hnonempty; hdiv n hn x hx (by rw [year, Nat.sub_self (15 ^ n)]; exact dvd_zero x)⟩
  · intro h; trivial
\end{lstlisting}
\end{minipage} \\
\hline

\end{longtable}

\end{document}